\PassOptionsToPackage{table,dvipsnames}{xcolor}
\documentclass{article} 
\usepackage{iclr2027_conference,times}

\usepackage{hyperref}
\usepackage{url}
\usepackage{capt-of}
\usepackage[utf8]{inputenc} 
\usepackage[T1]{fontenc}    
\usepackage{url}            
\usepackage{booktabs}       
\usepackage{amsfonts}       
\usepackage{nicefrac}       
\usepackage{microtype}      
\usepackage{amsmath}
\usepackage{multirow}
\usepackage{subcaption}
\usepackage{pifont}
\usepackage{algorithm}
\usepackage{bbm}

\usepackage{textcomp}
\usepackage{stfloats}
\usepackage{mathabx}
\usepackage{amsthm}
\usepackage{enumitem}
\setlist{leftmargin=5mm}
\usepackage{wrapfig}
\usepackage{makecell}
\usepackage{wrapfig}
\usepackage{lipsum}
\usepackage{amsmath, amssymb, graphicx, xcolor, xspace}
\usepackage{booktabs}
\usepackage{algpseudocode}
\usepackage{bbm}
\usepackage{tcolorbox}  
\usepackage{float}
\tcbuselibrary{breakable} 
\usepackage[utf8]{inputenc}
\usepackage{tcolorbox}
\usepackage{microtype}
\newtcolorbox{mybox}[1][]{
    title=#1,
    fonttitle=\small,
    fontupper=\small,
    left=1mm,
    right=1mm,
    top=1mm,
    bottom=0mm,
}
\definecolor{darkblue}{rgb}{0, 0, 0.5}
\definecolor{darkred}{rgb}{0.72, 0.22, 0.27}
\definecolor{lightblue}{RGB}{129, 209, 241}
\definecolor{forestgreen}{RGB}{34, 139, 34}
\hypersetup{
    colorlinks=true,
    citecolor=darkblue,
    linkcolor=darkblue,
    urlcolor=darkblue,
}

\usepackage{inconsolata}

\usepackage{graphicx}
\usepackage[utf8]{inputenc} 
\usepackage[T1]{fontenc}    
\usepackage{url}            
\usepackage{booktabs}       
\usepackage{amsfonts}       
\usepackage{nicefrac}       
\usepackage{microtype}      
\usepackage{xcolor}         
\usepackage{amssymb}
\usepackage{amsmath}
\usepackage{multirow}
\usepackage{subcaption}
\usepackage{pifont}
\usepackage{amssymb}
\usepackage{algorithm}
\usepackage{bbm}
\usepackage{stfloats}
\usepackage{mathabx}
\usepackage{amsthm}
\usepackage[table]{xcolor}
\usepackage{enumitem}
\setlist{leftmargin=5mm}
\usepackage{graphicx}
\usepackage{wrapfig}
\usepackage{makecell}
\usepackage[dvipsnames]{xcolor}
\usepackage{wrapfig}
\usepackage{lipsum}
\usepackage{amsmath, amssymb, graphicx, xcolor, xspace}
\usepackage{booktabs}
\usepackage{algpseudocode}
\usepackage{bbm}

\definecolor{gMMAU}{HTML}{C0D9ED}
\definecolor{gMMAR}{HTML}{F9E8C5}
\definecolor{gMMSU}{HTML}{CFBAD9}
\definecolor{gOVR}{HTML}{CFBAD9}
\usepackage{tikz}
\makeatletter
\newcommand{\DrawLine}{%
  \begin{tikzpicture}
  \path[use as bounding box] (0,0) -- (\linewidth,0);
  \draw[color=black,dashed,dash phase=2pt]
        (0-\kvtcb@leftlower-\kvtcb@boxsep,0)--
        (\linewidth+\kvtcb@rightlower+\kvtcb@boxsep,0);
  \end{tikzpicture}%
  }

\definecolor{violet}{RGB}{138, 43, 226}

\makeatother

\definecolor{citepcol}{HTML}{2DDC0E}
\definecolor{tableofcontent}{HTML}{E63E15}
\definecolor{urlcol}{HTML}{2470D8}
\definecolor{myorange}{RGB}{2, 142, 2}
\hypersetup{
    colorlinks=true,
    linkcolor=red,
    filecolor=magenta,      
    urlcolor=cyan,
}
\newcommand{\xhdr}[1]{{\vspace{1pt}\noindent\bfseries #1}.}

\definecolor{barGray}{HTML}{D9D9D9} 
\NewDocumentCommand{\shibo}{ mO{} }
{\textcolor{pink}{\textsuperscript{\textit{Shibo}}\textsf{\textbf{\small[#1]}}}}

\usepackage{hyperref}
\usepackage{fontawesome5}

\title{{\upshape\method}: Curiosity-Driven Test-Time Learning for Open-Ended Discovery}

\author{%
\textbf{Tao Feng}\textsuperscript{1},\;\;
\textbf{Fangxu Yu}\textsuperscript{2}, \;\;
\textbf{Zijie Lei}\textsuperscript{1}, \;\;
\textbf{Jiaru Zou}\textsuperscript{\textbf{3}},\;\;
\textbf{Changjiang Jiang}\textsuperscript{\textbf{1}},\;\;
\textbf{Yi Yan}\textsuperscript{\textbf{1}}, \\
\textbf{Jiaxuan You}\textsuperscript{\textbf{1}},\;\; 
\textbf{Pan Lu}\textsuperscript{\textbf{3}}\;\;
\\
\textsuperscript{1}University of Illinois Urbana-Champaign,\;\;
\textsuperscript{2}University of Maryland, College Park,\;\;\\
\textsuperscript{3}Stanford University
}

\newcommand{\method}{\texttt{\upshape CURIO}\xspace} 
\iclrfinalcopy 
\begin{document}

\maketitle

\begin{abstract}
Open-ended discovery requires learning from repeated attempts while continuing to explore directions whose value is not yet apparent. Search with a frozen large language model (LLM) can reuse previous solutions in context, but cannot update the model from its successes and failures on the test problem. Reinforcement learning (RL) enables such adaptation; however, strongly favoring high-reward trajectories may suppress low-reward yet potentially promising directions too early. We introduce \method, a curiosity-driven test-time learning framework that complements task feedback with an Intrinsic Curiosity World Model (ICWM). The ICWM learns transitions in the policy's hidden-state representation and supplies prediction-error bonuses at sampled tokens outside the policy's top-$k$ choices. Epoch normalization and an annealed weight regulate their contribution to the policy update. On six mathematical discovery tasks and single-cell denoising with Qwen3 backbones from 8B to 235B, three-run means improve over a matched task-only RL control on five mathematical objectives, match the best reported performance on Circle Packing, and improve denoising Score and mean squared error (MSE) on both held-out corpora at every tested scale. Relative gains reach 18.3\% on Hadamard and 10.8\% on denoising Score. Code-diversity measurements show greater structural variation among generated programs, supporting curiosity as a complementary exploration signal for learning in open-ended discovery.
\end{abstract}

\begin{center}
    \href{https://ft2023.github.io/CURIO/}{\textcolor{darkblue}{\faGlobe\ \texttt{Website}}}
    \quad
    \href{https://github.com/ft2023/CURIO}{\textcolor{darkblue}{\faGithub\ \texttt{Code}}}
\end{center}

\section{Introduction}
\label{sec:intro}

Open-ended scientific and algorithmic discovery seeks progressively better solutions rather than a single predetermined answer. LLMs can participate in this process by proposing programs, receiving evaluator feedback, and refining their attempts~\citep{romera2024mathematical,novikov2025alphaevolve}. This approach spans mathematical construction and broader AI R\&D tasks~\citep{tao2026mathematics,rank2026posttrainbench,wang2026naturebench,wu2026ai4ai}. Progress depends both on exploiting successful attempts and on exploring directions whose eventual value may not be reflected in their current reward.

Existing search-based auto-research methods repeatedly sample from a frozen LLM, organizing previous attempts through prompts, archives, and populations~\citep{novikov2025alphaevolve,lange2026shinkaevolve,ye2026structured}. These mechanisms can improve the context supplied to the model, but the model's parameters cannot learn from its successes and failures on the test problem. RL-based methods instead turn discovery into test-time RL, updating the proposing policy with experience collected from that same problem~\citep{wang2025thetaevolve,yuksekgonul2026learning}. This enables the model to internalize useful patterns as discovery proceeds.

\begin{wrapfigure}{R}{0.50\textwidth}
    \vspace{-4pt}
    \captionsetup{aboveskip=3pt}
    \centering
    \includegraphics[width=\linewidth]{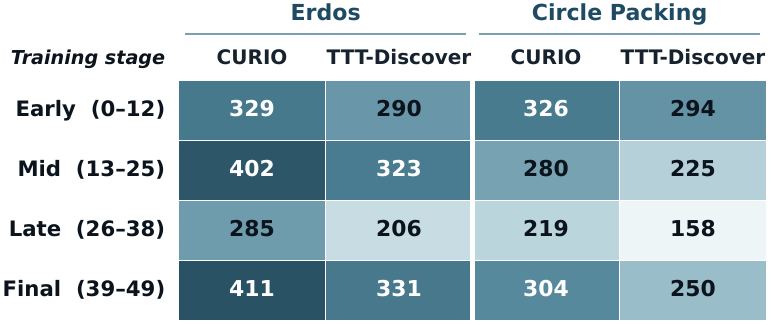}
    \caption{\textbf{Motivating exploration beyond task rewards.} On Erd\H{o}s and Circle Packing, \method has higher code diversity than task-only RL (TTT-Discover) at every displayed stage (metric: Appendix~\ref{app:code-diversity}).}
    \label{fig:modality_routing}
    \vspace{-10pt}
\end{wrapfigure}

Learning from task rewards nevertheless creates an exploration challenge. Updates that strongly prioritize high-reward trajectories can concentrate the policy around currently successful constructions. A direction with low current reward may still admit valuable refinements, yet receive too little future sampling to reveal that potential. Thus, improved exploitation can suppress promising alternatives prematurely. Our central question is: \textit{how can a policy learn from task feedback while retaining an incentive to explore underexplored directions?}

Task reward alone may provide insufficient incentive to revisit underexplored directions: it assesses the quality of a completed candidate rather than the potential of further exploration. Increasing sampling temperature broadens sampling, while entropy regularization encourages uncertainty without directly measuring learned transition predictability. We therefore seek an exploration signal learned from the policy's own experience and applied selectively during generation.

This requires resolving two linked questions. First, \textit{what experience should drive the exploration signal?} Surface differences between long code sequences need not imply different solution strategies. Policy hidden states instead provide a representation in which a learned model can assess transition predictability, although high prediction error does not by itself establish semantic novelty or future task improvement. Second, \textit{where should the signal affect learning?} Rewarding surprise at every token can distract from task progress. A selective rule and a controlled reward scale are needed to combine exploration with task-driven improvement.

We propose \method, a curiosity-driven test-time learning framework for open-ended discovery. Within the RL discovery loop, an ICWM learns token-level transitions from policy hidden-state features and sampled-token embeddings. Its cached prediction residuals supply an intrinsic signal, normalized over each epoch and retained only when a sampled token lies outside the policy's top-$k$ choices. An annealed weight combines this bonus with the task coefficient before the policy update. The ICWM models the policy's representation dynamics; the task evaluator continues to determine candidate quality. This supplies a complementary training signal without assuming that every surprising transition is useful or that every low-reward trajectory should be retained. Figure~\ref{fig:modality_routing} illustrates the resulting comparison: \method exhibits higher code-level structural diversity than the task-only RL baseline at every displayed stage on both tasks. This diagnostic does not by itself establish semantic diversity or a causal mechanism.

We evaluate \method on six mathematical discovery tasks and single-cell denoising on the held-out peripheral blood mononuclear cell (PBMC) and Tabula corpora, with Qwen3 backbones from 8B to 235B. Relative to a matched task-only RL control, three-run means improve on five mathematical objectives and match the best reported performance on Circle Packing; denoising Score and MSE improve on both corpora at every tested scale. Relative improvements reach 18.3\% on 235B Hadamard and 10.8\% on 8B Tabula Score. Figure~\ref{fig:code_solution_diversity} extends the introductory code-diversity comparison to all six discovery trajectories. Ablations examine curiosity strength, predictor architecture, and representation coupling. Together, these results evaluate the benefit of adding a learned exploration signal to reward-driven adaptation, while code diversity and prediction error remain distinct from semantic novelty.

\section{Preliminaries}
\label{sec:prelim}

\subsection{Open-Ended Discovery}
\label{sec:open-ended-discovery}
A discovery problem specifies a description $d$, candidate solutions, and an evaluator~\citep{novikov2025alphaevolve,yuksekgonul2026learning}. The aim is to find the best feasible solution within a finite budget; improving on a prior best-known value constitutes a new record. An archive $\mathcal{H}$ stores evaluated candidates and their programs. Given a parent $p\in\mathcal{H}$, the prompt $x=x(d,p)$ conditions a policy $\pi_\theta$, parameterized by $\theta$, to produce a completion $y$, containing code and optional reasoning. Parsing and executing the code yields a new candidate for evaluation and possible archive insertion. We write $R(x,y)$ for its scalar training reward after feasibility checks and the task-specific transformation in Appendix~\ref{app:tasks}; larger $R$ is better, even when the raw objective is minimized. The candidate $p$ and completion $y$ describe the discovery loop; $s_t$ and $a_t$ are reserved for the token-level ICWM quantities defined in \S\ref{sec:wm-curiosity}.

\subsection{RL for Discovery}
\label{sec:rl-discovery}
RL-based discovery learns from evaluator feedback on the current problem by updating the proposing policy~\citep{wang2025thetaevolve,yuksekgonul2026learning}. To emphasize high-reward candidates, a discovery-oriented objective uses exponential reward weighting~\citep{yuksekgonul2026learning}:
\begin{equation}
\label{eq:entropic}
J_\beta(\theta;x)=\log\mathbb{E}_{y\sim\pi_\theta(\cdot\mid x)}[e^{\beta R(x,y)}],\qquad \beta>0.
\end{equation}
Here $x$ is an archive-conditioned prompt, $R(x,y)$ is the evaluator reward, and $\beta$ controls the emphasis on high rewards. Larger $\beta$ more strongly favors high-reward completions, potentially reducing exploration of currently low-reward but promising directions.

\section{Curiosity-Driven Test-Time Learning}
\label{sec:method}
\begin{figure*}[t]
\centering
\includegraphics[width=1.0\textwidth]{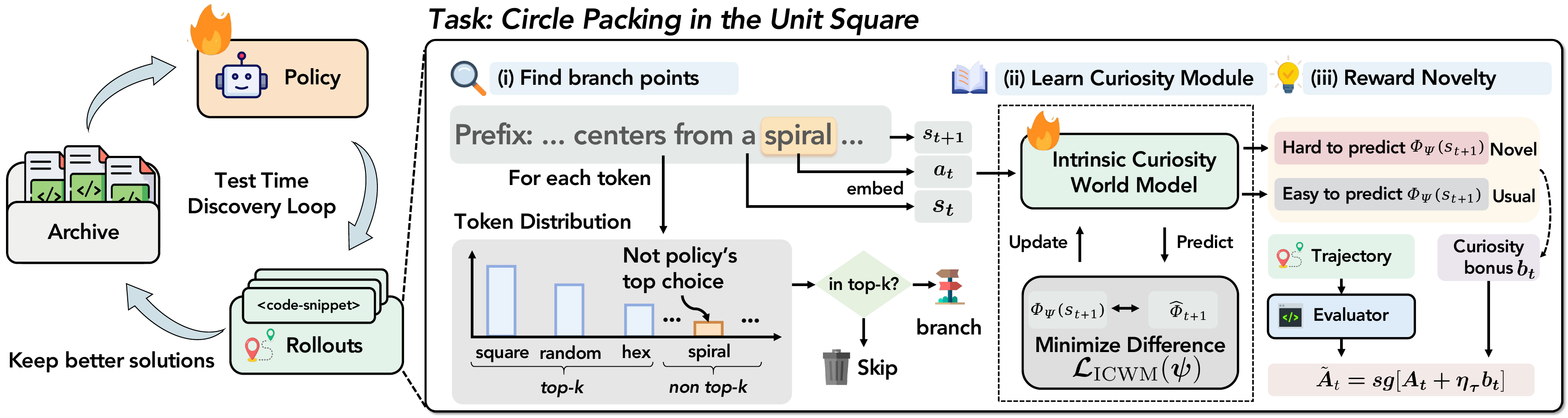}
\caption{\textbf{Overview of \method.} A policy expands archive parents into rollouts. The ICWM learns token-level transitions in encoded policy hidden states, and its prediction residual informs the curiosity bonus. The signal is retained at sampled tokens outside the policy's top-$k$ choices and combined with the task coefficient. The states $s_t,s_{t+1}$ and action $a_t$ are token-level features and embeddings. The diagram is schematic; Equations~\eqref{eq:sa}--\eqref{eq:final} specify the representations, normalization, and masking.}
\label{fig:step}
\end{figure*}

\method follows the RL discovery loop in \S\ref{sec:rl-discovery} and augments task-driven learning with a token-local curiosity signal. Figure~\ref{fig:step} summarizes the three components: select non-top-$k$ sampled tokens, learn an ICWM on policy representation transitions, and combine its prediction-error bonus with task feedback. We define the task-interface notation in \S\ref{sec:task-coefficients}, the ICWM and bonus in \S\ref{sec:wm-curiosity}, and the joint update in \S\ref{sec:joint-training}. Algorithm~\ref{alg:training} in Appendix~\ref{app:training} provides the complete procedure.

\xhdr{Overview}
In the diagram, $\Phi_\psi$ and $\widehat\Phi_{t+1}$ correspond to $\phi_\psi$ and $\hat\phi_{t+1}$ below, and the displayed $\eta_t$ means the epoch weight $\eta_\tau$. ``Branch points'' denotes the non-top-$k$ heuristic; ``novel'' and ``usual'' denote relative prediction difficulty, not established semantic novelty. ``Skip'' omits the curiosity bonus, not task learning or ICWM fitting. The bonus $b_t$ includes epoch normalization and masking; it is not the raw residual.

\subsection{Task Feedback and Token Notation}
\label{sec:task-coefficients}

\xhdr{Notation and token indexing}
At epoch $\tau$, $\theta_{\mathrm{old}}$ denotes the policy snapshot used for generation, and $\theta_0$ denotes the initial reference policy. Only adapter parameters are updated. A prompt $x$ has length $P\ge1$, its completion $y=(y_1,\ldots,y_T)$ has length $T$, and $z=(x,y)$ is indexed from zero. Thus $y_t$ occupies position $P+t-1$ and has conditioning prefix $z_{<P+t-1}$. The mask $m_t\in\{0,1\}$ is one for sampled, trainable completion tokens and zero for inserted forced-suffix tokens. Prompt and padding tokens are excluded from all completion-level sums. Let $\mathcal{D}_\tau$ be the epoch's training rollouts after filtering constant-reward groups, and let $M_\tau=\sum_{y\in\mathcal{D}_\tau}\sum_{t=1}^{T_y}m_t$. Here $T_y$ is the length of rollout $y$; each rollout retains its associated prompt $x$. Sums over $t$ run over its completion positions, and rollout indices on token quantities are suppressed. No training update is taken when $M_\tau=0$.

\xhdr{Sampled task coefficients}
We instantiate the exponential weighting of \S\ref{sec:rl-discovery} using the following construction, shared with the matched task-only control. The task coefficient supplies reward-driven learning; the ICWM adds a separate exploration signal.
For a group of $G\ge2$ rollouts from one parent, sampled from $\pi_{\theta_{\mathrm{old}}}(\cdot\mid x)$, with rewards $R_i=R(x,y^{(i)})$ for $i=1,\ldots,G$, the sampled task coefficient is
\begin{equation}
\label{eq:loo}
A_i^{\mathrm{task}}=\frac{e^{\beta R_i}}{\frac{1}{G-1}\sum_{j\ne i}e^{\beta R_j}}-1.
\end{equation}
The group-specific $\beta$ is treated as fixed during the policy update. Its calibration targets a Kullback--Leibler (KL) divergence of $\delta=\log 2$ between $\mathrm{softmax}(\beta(R_1,\ldots,R_G))$ and the uniform distribution over the $G$ rollouts when this target is attainable (Appendix~\ref{app:impl}). Equation~\eqref{eq:loo} is a finite-sample ratio estimator, not an unbiased estimate of the population-normalized exponential reward weight minus one. Only $A_i^{\mathrm{task}}$, before the token-local KL correction below, is shared by every token of rollout $i$.

\xhdr{Token-local reference correction}
For a token in rollout $i$, define the sampled log-ratio to the reference policy and its epoch mean:
\begin{align}
\label{eq:ref-logratio}
d_t&=\log\pi_{\theta_{\mathrm{old}}}(y_t\mid z_{<P+t-1})-\log\pi_{\theta_0}(y_t\mid z_{<P+t-1}),\\
\label{eq:kl}
\bar d&=\frac{1}{M_\tau}\sum_{y\in\mathcal{D}_\tau}\sum_t m_t d_t,
\qquad A_t=A_i^{\mathrm{task}}+\lambda m_t(\bar d-d_t).
\end{align}
Here, the reference-correction coefficient $\lambda\ge0$ controls a centered, token-local correction
toward the reference policy. This is the coefficient used by the training procedure; it is not a sequence-level log-ratio broadcast to every token. Set $d_t=0$ at inserted tokens, whose $m_t=0$ prevents any contribution. All sampled coefficients and reference statistics are held fixed during the policy update. This sampled, empirically centered correction should not be identified with the exact gradient of a sequence-level KL-regularized population objective.

\subsection{Intrinsic Curiosity World Model}
\label{sec:wm-curiosity}

The ICWM learns token-level transitions to supply an intrinsic exploration signal. It consists of a feature encoder $\phi_\psi$ and a forward predictor $f_\psi$ operating on actor hidden states and token embeddings. It models transitions within the policy's representation space, not the external scientific environment or the evaluator. Its prediction residual is converted into a token-local bonus through epoch normalization and selective masking.

\xhdr{ICWM states and actions}
Let $d_h$ be the backbone hidden dimension and $h_j\in\mathbb{R}^{d_h}$ be the last-layer hidden state at position $j$ from a no-gradient forward pass of $\pi_{\theta_{\mathrm{old}}}$. For completion step $t$, define
\begin{equation}
\label{eq:sa}
s_t=h_{P+t-2},\qquad a_t=\mathrm{Emb}(y_t),\qquad s_{t+1}=h_{P+t-1}.
\end{equation}
Here $\mathrm{Emb}$ is the frozen backbone token-embedding lookup, so $a_t\in\mathbb{R}^{d_h}$. In particular, $s_1=h_{P-1}$ is the final prompt representation. The symbols $s_t,a_t$ refer to token-level features and embeddings, whereas $p$ denotes an archive candidate and $y$ a complete rollout. A single hidden state is a feature representation, not an assertion that it contains the full Transformer attention cache.

\xhdr{ICWM learning and prediction residuals}
Following the prediction-error approach to curiosity~\citep{pathak2017curiosity}, an encoder $\phi_\psi:\mathbb{R}^{d_h}\to\mathbb{R}^{d_h}$ and predictor $f_\psi:\mathbb{R}^{2d_h}\to\mathbb{R}^{d_h}$ define
\begin{equation}
\label{eq:icm}
\hat\phi_{t+1}=f_\psi([\phi_\psi(s_t);a_t]),\qquad
 e_t(\psi)=\tfrac12\|\phi_\psi(s_{t+1})-\hat\phi_{t+1}\|_2.
\end{equation}
Here $[\cdot;\cdot]$ denotes concatenation, and $\psi$ collects the parameters of both networks. The encoder and predictor are jointly fitted as the ICWM on mini-batches $\mathcal{B}\subset\mathcal{D}_\tau$ using
\begin{equation}
\label{eq:icmloss}
\mathcal{L}_{\mathrm{ICWM}}(\psi;\mathcal{B})=
\frac{\sum_{y\in\mathcal{B}}\sum_t m_t e_t(\psi)}{\sum_{y\in\mathcal{B}}\sum_t m_t}.
\end{equation}
The norm in Equation~\eqref{eq:icm} is the Euclidean norm, not its square. For mini-batches with no trainable tokens, set the cached residuals to zero and skip the ICWM update. Otherwise, for each mini-batch, we cache $e_t=\mathrm{sg}[e_t(\psi)]$ before its optimizer step, then update $\psi$; $\mathrm{sg}$ denotes stop-gradient. Thus, the cached residuals are online prediction errors evaluated at successive ICWM parameter states, not a rescore of the epoch using one final predictor. No ICWM gradient is propagated into the actor.

The code-diversity diagnostic in Figure~\ref{fig:modality_routing} compares generated code strings (Appendix~\ref{app:code-diversity}); it is not the ICWM residual $e_t$ or bonus $b_t$, and does not enter the training objective. High ICWM prediction error identifies transitions that the current predictor models poorly. It does not establish semantic novelty, insufficient prior exploration, or future task improvement, and also depends on predictor capacity and representation drift. Moreover, joint training of the encoder and predictor with this loss alone admits constant-feature solutions. The formulation does not provide an anti-collapse guarantee; this limitation should be distinguished from the empirical task performance.

\xhdr{Epoch normalization}
Let $\mathcal{I}_\tau=\{(y,t):y\in\mathcal{D}_\tau,1\le t\le T_y\}$ be the set of actual completion positions, including inserted suffix positions but excluding prompts and padding. We define
\begin{align}
\label{eq:epoch-stats}
\mu_\tau&=\frac{1}{|\mathcal{I}_\tau|}\sum_{(y,t)\in\mathcal{I}_\tau}m_t e_t,
&\sigma_\tau^2&=\frac{1}{|\mathcal{I}_\tau|}\sum_{(y,t)\in\mathcal{I}_\tau}(m_t e_t-\mu_\tau)^2,\\
\label{eq:whiten}
\hat e_t&=\mathrm{sg}\!\left[\frac{m_t e_t-\mu_\tau}{\sqrt{\sigma_\tau^2+\varepsilon}}\right],&&\varepsilon>0.
\end{align}
Here $\mu_\tau$ and $\sigma_\tau^2$ are the epoch mean and variance of the masked residuals, and $\varepsilon$ is a positive numerical stabilizer. This convention treats inserted tokens as zero residuals in the epoch statistics; it is not normalization over trainable tokens only. It centers the signal over $\mathcal{I}_\tau$ and rescales prediction errors, but does not ensure that their scale matches the task coefficients or that a subsequently masked signal remains centered.

\xhdr{Selective curiosity bonus}
We retain the normalized residual only at sampled tokens outside the sampling policy's top-$k$ choices:
\begin{equation}
\label{eq:mask}
u_t=m_t\,\mathbf{1}\!\left[y_t\notin\mathrm{TopK}_k\!\left(\pi_{\theta_{\mathrm{old}}}(\cdot\mid z_{<P+t-1})\right)\right],
\end{equation}
The selected tokens receive the normalized prediction-error bonus
\begin{equation}
\label{eq:bonus}
b_t=u_t\hat e_t.
\end{equation}
Here $\mathbf{1}[\cdot]$ is the indicator function, and $\mathrm{TopK}_k$ returns the $k$ highest-probability token identities. With $k=1$, this selects departures from the highest-probability token, using the inference engine's tie-breaking convention. It is a heuristic for directing exploration, not an entropy threshold or a guarantee that selected positions are semantic branch points. The mask is recorded during generation and held fixed during training. In general, $\mathbb{E}[b_t]\ne0$ even when $\mathbb{E}[\hat e_t]=0$; we therefore call $b_t$ a token-local bonus, not an unbiased intrinsic advantage. It is applied at the current token without accumulating future intrinsic rewards. Because normalization is centered, $b_t$ can be positive or negative at selected tokens; unselected tokens still receive task feedback and can train the ICWM through $m_t$.

\subsection{Joint Optimization and Training Procedure}
\label{sec:joint-training}

\xhdr{Joint policy update}
The update combines the task coefficient with the bonus in Equation~\eqref{eq:bonus}:
\begin{equation}
\label{eq:final}
\tilde A_t=\mathrm{sg}[A_t+\eta_\tau b_t].
\end{equation}
The curiosity weight follows
\begin{equation}
\label{eq:eta}
\eta_\tau=\eta_0\max\!\left(\alpha,\tfrac12[1+\cos(\pi\tau/E)]\right).
\end{equation}
Here $E$ is the total number of training epochs, $\tau=0,\ldots,E-1$ indexes epochs, $\eta_0\ge0$ is the initial weight, and $0<\alpha\le1$ is the relative floor. The coefficient is annealed toward its floor over training. Starting from $\theta=\theta_{\mathrm{old}}$, one adapter optimizer step minimizes
\begin{equation}
\label{eq:pg}
\mathcal{L}_{\mathrm{PG}}(\theta)=
-\sum_{y\in\mathcal{D}_\tau}\sum_{t=1}^{T_y}m_t
\frac{\pi_\theta(y_t\mid z_{<P+t-1})}{\pi_{\theta_{\mathrm{old}}}(y_t\mid z_{<P+t-1})}\,\tilde A_t.
\end{equation}
The denominator and $\tilde A_t$ are detached; only the numerator is differentiated. This is a token-level importance-weighted surrogate evaluated on the sampled batch. At $\theta=\theta_{\mathrm{old}}$, its gradient is the negative weighted token score-function sum. Finite-group weighting, group filtering, the sampled reference correction, and the token-local bonus distinguish this update from the exact gradient of the objective in Equation~\eqref{eq:entropic}. Setting $\eta_0=0$ removes the curiosity bonus and recovers the task-only control update. Constant-reward groups are filtered before either loss, so they contribute no curiosity update either.

\xhdr{End-to-end training procedure}
Algorithm~\ref{alg:training} (Appendix~\ref{app:training}) summarizes each epoch. We collect and evaluate rollouts with $\theta_{\mathrm{old}}$, update the archive, and filter constant-reward groups. Retained rollouts receive task coefficients from Equations~\eqref{eq:loo}--\eqref{eq:kl}. When curiosity is enabled, we cache residuals before each ICWM update (Equations~\eqref{eq:icm}--\eqref{eq:icmloss}), then normalize and mask them (Equations~\eqref{eq:epoch-stats}--\eqref{eq:bonus}). Finally, Equations~\eqref{eq:final}--\eqref{eq:pg} combine the signals for one adapter update; the actor remains fixed until this step.

\section{Experiments}
\label{sec:exp}

\subsection{Experimental Setup}
\label{subsec:setup}

\xhdr{Tasks and benchmarks}
We evaluate on two task families. For \textbf{mathematical discovery}, following \citet{yuksekgonul2026learning} and \citet{novikov2025alphaevolve}, we consider six open problems: Erd\H{o}s minimum-overlap problem (\textbf{Erd\H{o}s}, lower is better), three autocorrelation inequalities (\textbf{AC1}--\textbf{AC3}), the \textbf{Circle Packing} problem with $n{=}26$ circles in the unit square (sum of radii, higher is better; best-known $\approx 2.635983$), and the \textbf{HadamardMatrix} task following \citet{wang2025thetaevolve}, with task settings specified in Appendix~\ref{app:tasks}. For \textbf{single-cell denoising}, we adopt the biology task of \citet{yuksekgonul2026learning}, in which the policy evolves a denoising program evaluated on two single-cell RNA sequencing (scRNA-seq) corpora, \textbf{PBMC} and \textbf{Tabula}, under three complementary criteria: a benchmark Score (higher is better), reconstruction MSE (lower is better), and Poisson negative log-likelihood (NLL; lower is better).

\xhdr{Baselines}
We compare two baseline families: \textbf{frozen-LLM search}, comprising \textsc{AlphaEvolve}~\citep{novikov2025alphaevolve} via \textsc{OpenEvolve}~\citep{openevolve}, \textsc{ShinkaEvolve}~\citep{lange2026shinkaevolve}, \textsc{Best-of-25600}, and \textsc{SimpleTES}~\citep{ye2026structured}; and \textbf{trainable test-time RL}, comprising \textsc{ThetaEvolve}~\citep{wang2025thetaevolve} and \textsc{TTT-Discover}~\citep{yuksekgonul2026learning}. Frozen baselines vary candidates, prompts, and histories while keeping policy parameters fixed; SimpleTES uses independent trajectories, iterative refinement, local selection, and history reuse under $25{,}600$ nominal generation slots. ThetaEvolve always denotes its test-time RL variant; both RL baselines update the policy from task feedback. TTT-Discover is the matched task-only control, using \S\ref{sec:method} with $\eta_0=0$, while \method adds ICWM-derived curiosity to test-time RL, comparing exploration among learning policies. We use Qwen3 backbones at 8B, 14B, and 235B~\citep{yang2025qwen3}; SimpleTES is evaluated at 8B and 14B, and the 235B comparison includes only TTT-Discover and \method. Budgets, optimization settings, and implementation details are provided in Appendix~\ref{app:impl}.

\xhdr{Evaluation protocol}
For mathematical discovery, we report the best feasible raw objective found during each run, minimized for Erd\H{o}s, AC1, and AC3, and maximized for AC2, Circle Packing, and Hadamard. For denoising, the epoch is selected by held-in MSE, and that epoch's Score, MSE, and Poisson NLL on PBMC and Tabula are reported together; neither held-out corpus is used for epoch selection (Appendix~\ref{app:task-denoising}). For every method, backbone, and task configuration in Tables~\ref{tab:discovery} and~\ref{tab:singlecell}, we report the arithmetic mean over three runs. Mathematical objectives are averaged after selecting the best feasible value within each run; denoising metrics are averaged after the within-run checkpoint selection described above. The main tables show means only for space; mean differences alone do not establish statistical significance. Prompts and evaluator specifications are in Appendix~\ref{app:tasks} and~\ref{app:prompt}.

\subsection{Task Performance and Code Diversity}

Tables~\ref{tab:discovery} and~\ref{tab:singlecell} summarize the reported results.
\textbf{(i) Comparison with the task-only control.} \method improves on TTT-Discover for Erd\H{o}s, AC1--AC3, and Hadamard at each tested scale, and matches the best reported performance on Circle Packing at the displayed precision. It also improves denoising Score and MSE on both held-out corpora. These are not uniform wins over every baseline: on 8B Tabula, ThetaEvolve has higher Score ($0.7207$ versus $0.7165$) and lower MSE ($0.1441$ versus $0.1485$).

\textbf{(ii) Dependence on model scale.} The Hadamard relative improvement over TTT-Discover is $4.11\%$, $3.36\%$, and $18.33\%$ at 8B, 14B, and 235B, respectively. The largest observed advantage is therefore at 235B, but the trend is not monotonic. The absolute Erd\H{o}s objective improves from $0.380932$ to $0.380919$ to $0.380880$ as model size increases; this does not imply a growing relative advantage over the baseline. Attributing either trend specifically to richer hidden representations would require additional controls.

\begin{figure*}[!t]
    \centering
    \begin{subfigure}[t]{0.32\textwidth}
        \centering
        \includegraphics[width=\linewidth]{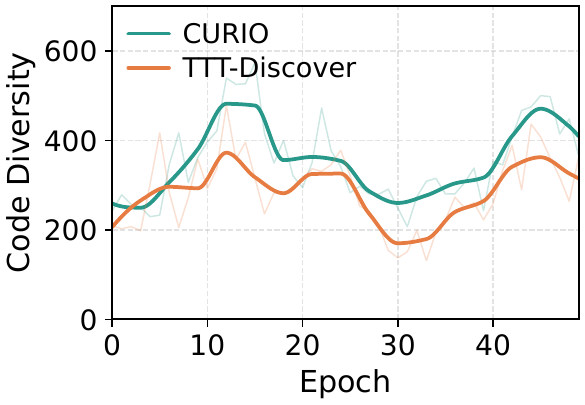}
        \caption{Erd\H{o}s}
        \label{fig:diversity-erdos}
    \end{subfigure}\hfill
    \begin{subfigure}[t]{0.32\textwidth}
        \centering
        \includegraphics[width=\linewidth]{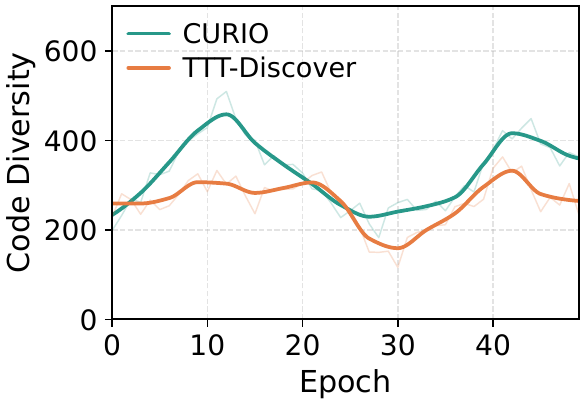}
        \caption{AC1}
        \label{fig:diversity-ac1}
    \end{subfigure}\hfill
    \begin{subfigure}[t]{0.32\textwidth}
        \centering
        \includegraphics[width=\linewidth]{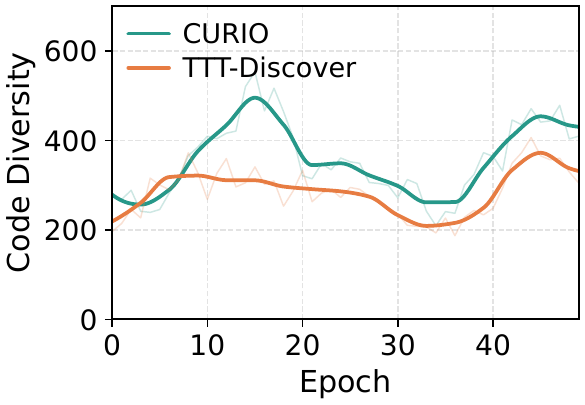}
        \caption{AC2}
        \label{fig:diversity-ac2}
    \end{subfigure}\par\vspace{4pt}
    \begin{subfigure}[t]{0.32\textwidth}
        \centering
        \includegraphics[width=\linewidth]{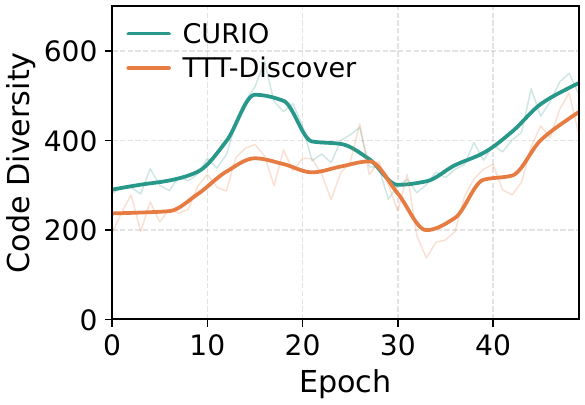}
        \caption{AC3}
        \label{fig:diversity-ac3}
    \end{subfigure}\hfill
    \begin{subfigure}[t]{0.32\textwidth}
        \centering
        \includegraphics[width=\linewidth]{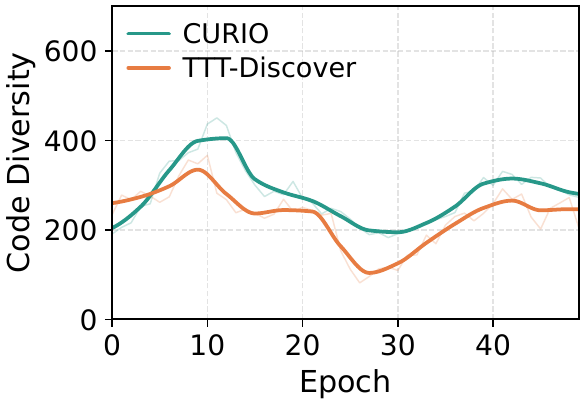}
        \caption{Circle Packing}
        \label{fig:diversity-circle_packing}
    \end{subfigure}\hfill
    \begin{subfigure}[t]{0.32\textwidth}
        \centering
        \includegraphics[width=\linewidth]{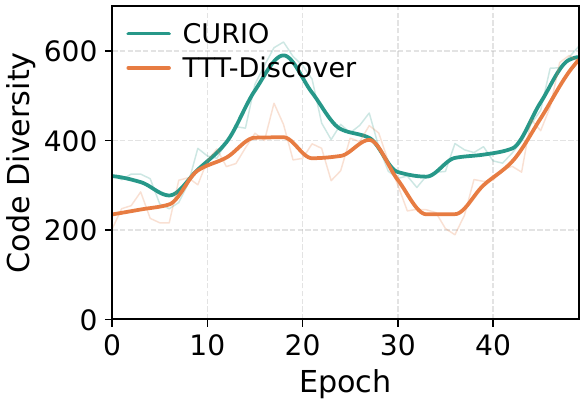}
        \caption{HadamardMatrix}
        \label{fig:diversity-hadamardmatrix}
    \end{subfigure}
    \caption{\textbf{Code-level solution diversity across six discovery tasks.} Panels are ordered left to right, top to bottom: Erd\H{o}s, AC1, AC2, AC3, Circle Packing, and HadamardMatrix. Each panel displays measured code diversity $D$ over epochs for \method and TTT-Discover, using OpenEvolve's island code diversity metric~\citep{openevolve} (Appendix~\ref{app:code-diversity}); higher $D$ indicates greater code-level structural variation.}
    \label{fig:code_solution_diversity}
\end{figure*}

\textbf{(iii) Code-level solution diversity.} We measure code-level structural variation using OpenEvolve's \textit{island code diversity metric}~\citep{openevolve}, which compares pairs of generated programs through differences in code length, line count, and character sets. Appendix~\ref{app:code-diversity} defines the pairwise score and its aggregation. Figure~\ref{fig:code_solution_diversity} shows the measured trajectories across six mathematical discovery tasks. \method generally maintains higher code diversity $D$ than TTT-Discover over the displayed epochs, although the curves cross or approach one another at some stages. Figure~\ref{fig:modality_routing} summarizes the same code-diversity diagnostic by training stage for Erd\H{o}s and Circle Packing. Across the early (epochs 0--12), middle (13--25), late (26--38), and final (39--49) stages, CURIO has higher displayed $D$ than TTT-Discover on both tasks. These observations support greater code-level structural diversity, but do not establish semantic diversity of the underlying constructions, unchanged token entropy, or the isolated contribution of non-top-$k$ masking.

\begin{table}[t]
\centering
\caption{
\textbf{Test-time discovery across mathematical and combinatorial tasks.}
Results are grouped by policy-backbone scale.
All results are means over three runs; variability is omitted for space.
Arrows denote improvement direction.
Within each scale block, \textbf{best} is bold and \underline{second-best} is underlined.
Ties at the displayed precision share the same marking.
\method improves over TTT-Discover on five objectives and matches the best reported performance on Circle Packing across model scales,
with particularly strong gains on HadamardMatrix.
}
\label{tab:discovery}

\setlength{\tabcolsep}{8pt}
\renewcommand{\arraystretch}{1.28}

\resizebox{0.98\textwidth}{!}{%
\begin{tabular}{l l cccccc}
\toprule
Scale & Method
& Erd\H{o}s $(\downarrow)$
& AC1 $(\downarrow)$
& AC2 $(\uparrow)$
& AC3 $(\downarrow)$
& Circle Packing $(\uparrow)$
& HadamardMatrix $(\uparrow)$ \\
\midrule

\multirow{7}{*}{8B}
 & \textsc{AlphaEvolve}
 & 0.381002
 & 1.50561
 & 0.8947
 & 1.5062
 & 2.63585
 & 0.5589 \\

 & \textsc{ThetaEvolve}
 & 0.380999
 & 1.50649
 & 0.9449
 & 1.4942
 & \underline{2.63597}
 & 0.5748 \\

 & \textsc{ShinkaEvolve}
 & 0.381026
 & 1.51493
 & 0.8501
 & 1.5074
 & \underline{2.63597}
 & 0.5661 \\

 & \textsc{Best-of-25600}
 & 0.381091
 & 1.51032
 & 0.9318
 & 1.5081
 & 2.63570
 & 0.5554 \\

 & \textsc{SimpleTES}
 & 0.381006
 & 1.50628
 & 0.9458
 & 1.4951
 & \underline{2.63597}
 & 0.5776 \\

 & \textsc{TTT-Discover}
 & \underline{0.380995}
 & \underline{1.50546}
 & \underline{0.9463}
 & \underline{1.4935}
 & \textbf{2.63598}
 & \underline{0.5809} \\

 & \method
 & \textbf{0.380932}
 & \textbf{1.50513}
 & \textbf{0.9481}
 & \textbf{1.4846}
 & \textbf{2.63598}
 & \textbf{0.6048} \\

\midrule

\multirow{7}{*}{14B}
 & \textsc{AlphaEvolve}
 & 0.381068
 & 1.50971
 & 0.9158
 & 1.5109
 & 2.63482
 & 0.5762 \\

 & \textsc{ThetaEvolve}
 & 0.383710
 & 1.51488
 & 0.9351
 & 1.6428
 & 2.62491
 & 0.5867 \\

 & \textsc{ShinkaEvolve}
 & 0.381041
 & 1.51052
 & 0.9048
 & 1.4996
 & 2.63509
 & 0.5826 \\

 & \textsc{Best-of-25600}
 & 0.381119
 & 1.50997
 & 0.9271
 & 1.5051
 & 2.63439
 & 0.5708 \\

 & \textsc{SimpleTES}
 & 0.381009
 & 1.50819
 & 0.9374
 & 1.4879
 & \underline{2.63546}
 & \underline{0.5902} \\

 & \textsc{TTT-Discover}
 & \underline{0.380979}
 & \underline{1.50608}
 & \underline{0.9391}
 & \underline{1.4592}
 & \textbf{2.63598}
 & 0.5864 \\

 & \method
 & \textbf{0.380919}
 & \textbf{1.50591}
 & \textbf{0.9418}
 & \textbf{1.4578}
 & \textbf{2.63598}
 & \textbf{0.6061} \\

\midrule

\multirow{2}{*}{235B}
& \textsc{TTT-Discover}
 & \underline{0.380918}
 & \underline{1.50376}
 & \underline{0.9562}
 & \underline{1.4561}
 & \textbf{2.63598}
 & \underline{0.7846} \\

 & \method
 & \textbf{0.380880}
 & \textbf{1.50315}
 & \textbf{0.9588}
 & \textbf{1.4549}
 & \textbf{2.63598}
 & \textbf{0.9284} \\

\bottomrule
\end{tabular}%
}

\end{table}

\begin{table}[t]
\centering
\caption{
\textbf{Single-cell denoising.}
Results on PBMC and Tabula scRNA-seq corpora, grouped by policy-backbone scale.
All results are means over three runs; variability is omitted for space.
Arrows denote improvement direction.
Within each scale block, \textbf{best} is bold and \underline{second-best} is underlined.
Ties at the displayed precision share the same marking.
\method consistently improves over \textsc{TTT-Discover} across model scales.
}
\label{tab:singlecell}

\setlength{\tabcolsep}{8pt}
\renewcommand{\arraystretch}{1.25}

\resizebox{\textwidth}{!}{%
\begin{tabular}{l l cccccc}
\toprule
\multirow{2}{*}{Scale}
& \multirow{2}{*}{Method}
& \multicolumn{3}{c}{PBMC}
& \multicolumn{3}{c}{Tabula} \\
\cmidrule(lr){3-5}
\cmidrule(lr){6-8}

&
& Score $(\uparrow)$
& MSE $(\downarrow)$
& Poisson $(\downarrow)$
& Score $(\uparrow)$
& MSE $(\downarrow)$
& Poisson $(\downarrow)$ \\

\midrule

\multirow{7}{*}{8B}
 & \textsc{AlphaEvolve}
 & 0.6738
 & 0.1817
 & 0.0554
 & 0.6785
 & 0.1716
 & 0.0323 \\

 & \textsc{ThetaEvolve}
 & \underline{0.6981}
 & \underline{0.1588}
 & \underline{0.0494}
 & \textbf{0.7207}
 & \textbf{0.1441}
 & \underline{0.0302} \\

 & \textsc{ShinkaEvolve}
 & 0.6787
 & 0.1764
 & 0.0535
 & 0.6881
 & 0.1669
 & 0.0312 \\

 & \textsc{Best-of-25600}
 & 0.6274
 & 0.2071
 & 0.0607
 & 0.6382
 & 0.1968
 & 0.0354 \\

 & \textsc{SimpleTES}
 & 0.6958
 & 0.1641
 & 0.0503
 & 0.7118
 & 0.1514
 & 0.0305 \\

 & \textsc{TTT-Discover}
 & 0.6842
 & 0.2087
 & 0.0818
 & 0.6468
 & 0.1963
 & 0.0363 \\

 & \method
 & \textbf{0.7009}
 & \textbf{0.1552}
 & \textbf{0.0493}
 & \underline{0.7165}
 & \underline{0.1485}
 & \textbf{0.0301} \\

\midrule

\multirow{7}{*}{14B}
 & \textsc{AlphaEvolve}
 & 0.6684
 & 0.1821
 & 0.0504
 & 0.6586
 & 0.1824
 & 0.0302 \\

 & \textsc{ThetaEvolve}
 & 0.6814
 & 0.1678
 & 0.0503
 & 0.6892
 & 0.1586
 & 0.0302 \\

 & \textsc{ShinkaEvolve}
 & 0.6782
 & 0.1715
 & 0.0502
 & 0.6783
 & 0.1618
 & \underline{0.0301} \\

 & \textsc{Best-of-25600}
 & 0.6171
 & 0.2023
 & 0.0506
 & 0.6576
 & 0.1822
 & 0.0303 \\

 & \textsc{SimpleTES}
 & 0.6873
 & 0.1651
 & 0.0499
 & 0.6966
 & 0.1558
 & 0.0302 \\

 & \textsc{TTT-Discover}
 & \underline{0.6901}
 & \underline{0.1615}
 & \underline{0.0492}
 & \underline{0.7012}
 & \underline{0.1526}
 & \underline{0.0301} \\

 & \method
 & \textbf{0.7052}
 & \textbf{0.1555}
 & \textbf{0.0491}
 & \textbf{0.7218}
 & \textbf{0.1423}
 & \textbf{0.0292} \\

\midrule

\multirow{2}{*}{235B}
& \textsc{TTT-Discover}
 & \underline{0.7011}
 & \underline{0.1562}
 & \underline{0.0491}
 & \underline{0.7254}
 & \underline{0.1401}
 & \underline{0.0291} \\

 & \method
 & \textbf{0.7156}
 & \textbf{0.1501}
 & \textbf{0.0490}
 & \textbf{0.7408}
 & \textbf{0.1334}
 & \textbf{0.0281} \\

\bottomrule
\end{tabular}%
}

\end{table}

\captionsetup[subfigure]{justification=centering}
\begin{figure*}[t]
    \centering

    \begin{subfigure}[t]{0.32\textwidth}
        \centering
        \includegraphics[height=4.4cm]{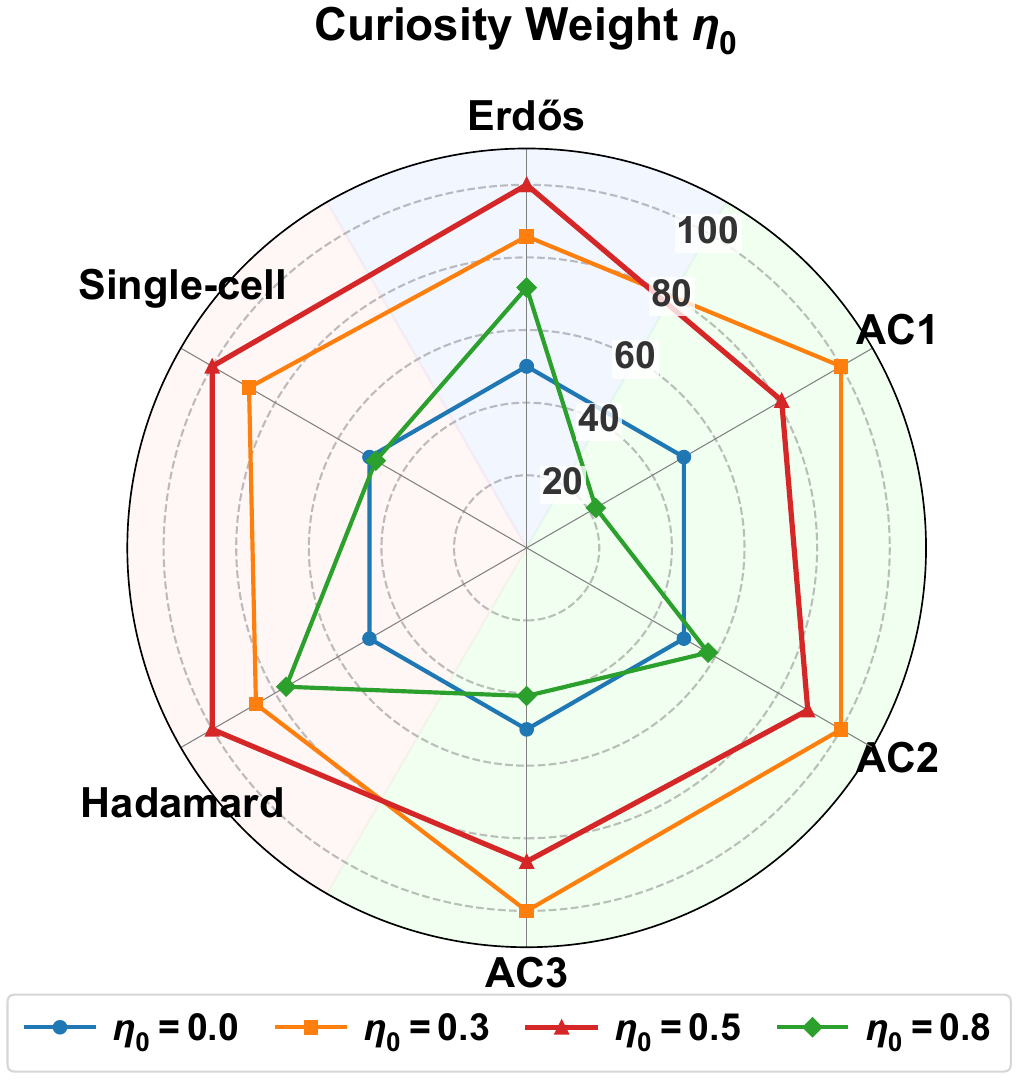}
        \caption*{(a)}
        \label{fig:ablation_curiosity_weight}
    \end{subfigure}
    %
    \begin{subfigure}[t]{0.32\textwidth}
        \centering
        \includegraphics[height=4.4cm]{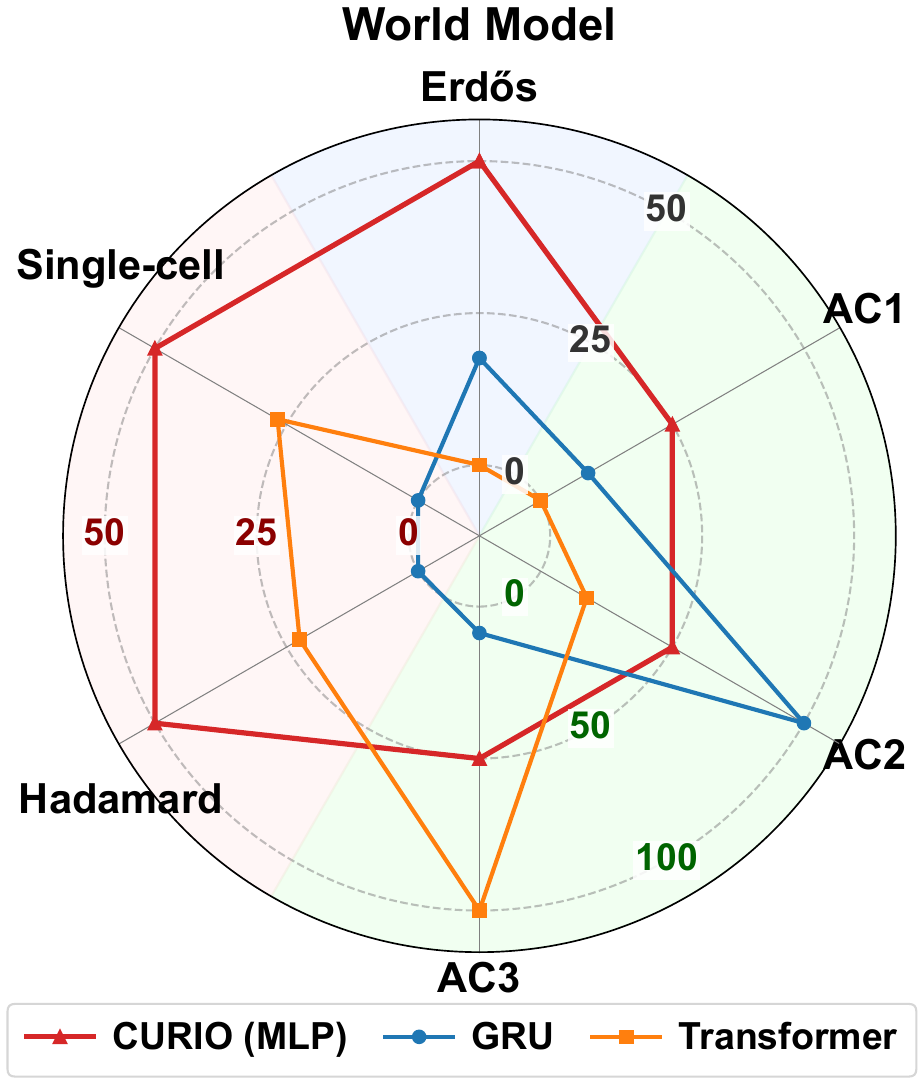}
        \caption*{(b)}
        \label{fig:ablation_world_model}
    \end{subfigure}
    %
    \begin{subfigure}[t]{0.32\textwidth}
        \centering
        \includegraphics[height=4.4cm]{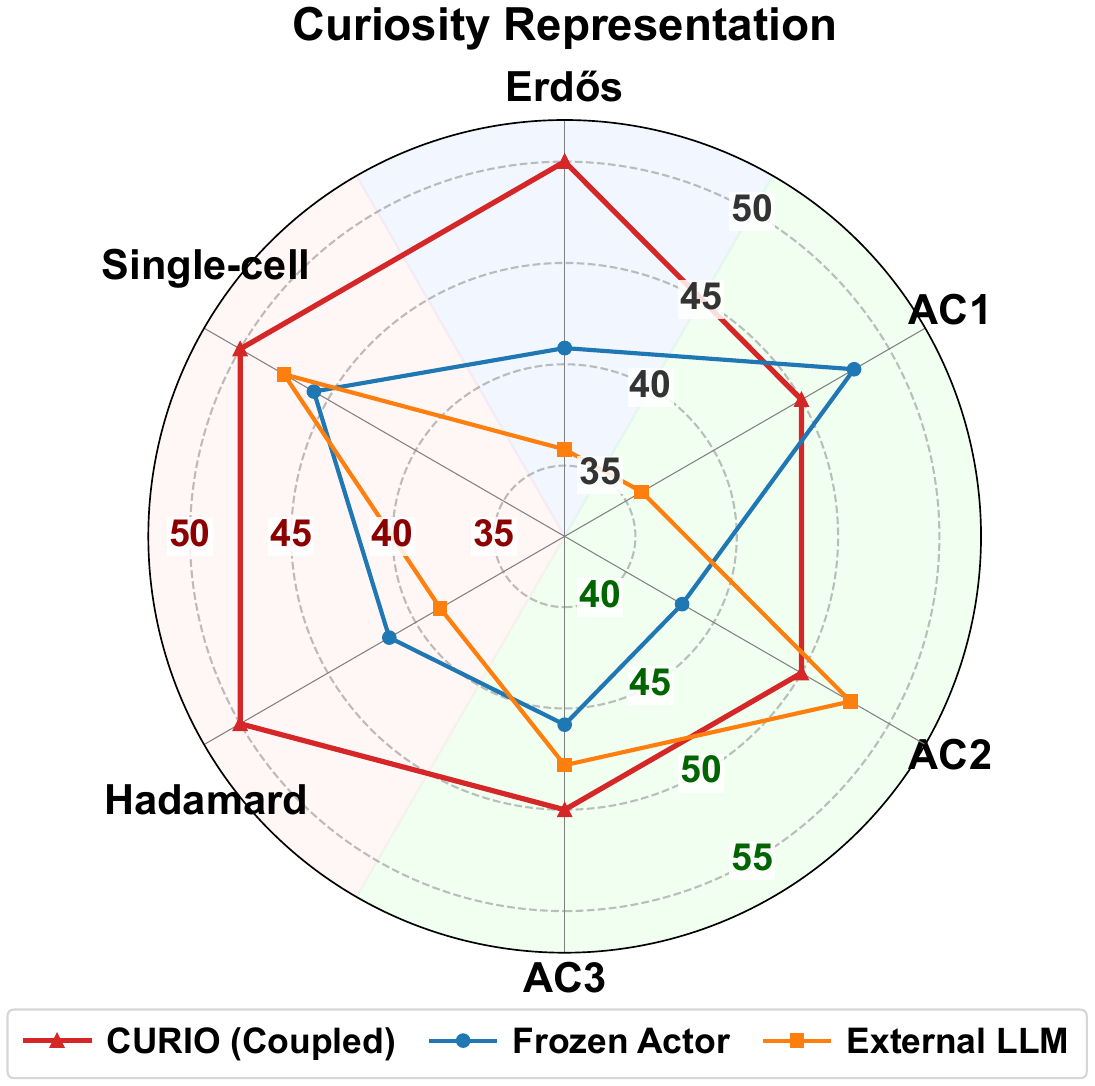}
        \caption*{(c)}
        \label{fig:ablation_curiosity_representation}
    \end{subfigure}

    \caption{
    \textbf{Ablation studies of \method on the Qwen3-14B backbone.}
    We report direction-aware normalized performance across six discovery tasks, with larger values indicating better performance.
    For each panel, the reference configuration is fixed at 50 on every task, so values above (below) 50 indicate improvement (degradation) relative to the reference.
    (a) \textit{Curiosity weight}: we vary the initial curiosity weight $\eta_0 \in \{0, 0.3, 0.5, 0.8\}$, using $\eta_0=0$ (TTT-DISCOVER) as the reference.
    (b) \textit{ICWM architecture}: we compare the ICWM's default multilayer perceptron (MLP) forward predictor with gated recurrent unit (GRU) and Transformer variants, using \method~(MLP) as the reference.
    (c) \textit{Curiosity representation}: we compare the ICWM's policy-coupled input representation with a frozen-actor representation and a frozen external-LLM representation, using \method~(Coupled) as the reference.
    Single-cell averages the PBMC and Tabula results.
    }
    \label{fig:method_ablation_radar}
\end{figure*}

\subsection{Ablation Studies of the ICWM}
\label{sec:ablation}

\vspace{-1.8mm}

To examine the sensitivity of the key design choices in \method, we conduct ablation studies on the Qwen3-14B backbone across six discovery tasks: Erd\H{o}s, AC1, AC2, AC3, Hadamard, and single-cell denoising.
We study three aspects of the ICWM: the bonus weight, the architecture of the ICWM's forward predictor, and the input representation used to model token-level transitions.
Results are reported in Figure~\ref{fig:method_ablation_radar}.

\begin{itemize}[leftmargin=1em, itemsep=0pt, topsep=0pt, parsep=0pt, partopsep=0pt]
    \item \textbf{Curiosity Weight}: We vary the initial curiosity weight $\eta_0 \in \{0, 0.3, 0.5, 0.8\}$, where $\eta_0=0$ removes the curiosity reward and recovers TTT-DISCOVER. This evaluates the trade-off between task exploitation and curiosity-driven exploration.

    \item \textbf{ICWM Architecture}: We replace the ICWM's default MLP forward predictor with GRU- and Transformer-based predictors while keeping the remaining training procedure unchanged. This evaluates whether the gains depend on the predictor architecture within the ICWM.

    \item \textbf{Curiosity Representation}: We compare the ICWM's policy-coupled input representation with two decoupled alternatives: a frozen representation from the initial actor and a frozen representation from an independent external LLM. This evaluates whether the prediction-based curiosity signal benefits from modeling representations of the evolving policy itself.
\end{itemize}

Figure~\ref{fig:method_ablation_radar} indicates sensitivity to curiosity strength. Among the plotted settings, $\eta_0=0.3$ performs best on the autocorrelation tasks, while $\eta_0=0.5$ is preferable on Erd\H{o}s, Hadamard, and the aggregated single-cell task. The largest tested weight, $\eta_0=0.8$, underperforms the task-only control on some tasks. Appendix~\ref{app:impl} lists the task-specific weights selected from this sweep and the default for Circle Packing. The MLP reference is competitive with the recurrent and Transformer alternatives, but the best architecture depends on the task. Likewise, the coupled representation leads on most plotted tasks, with exceptions for frozen representations on individual tasks. These comparisons examine predictor and representation choices; they do not independently validate token-mask selection, rule out encoder collapse, or establish stability across random seeds.

\section{Additional Related Work}
LLM-based discovery combines program generation with automated evaluation and population search~\citep{lehman2023evolution,meyerson2024language,romera2024mathematical,novikov2025alphaevolve,openevolve,lange2026shinkaevolve,assumpccao2025codeevolve}, with applications to reward design, equation discovery, GPU kernels, and agentic systems~\citep{ma2024eureka,shojaee2025llm,ouyang2025kernelbench,hu2025automated}. Test-time RL additionally learns from experience on the current problem~\citep{wang2025thetaevolve,yuksekgonul2026learning}; our ThetaEvolve and TTT-Discover baselines update their policies, whereas our SimpleTES baseline uses frozen-policy search with parallel trajectories, sequential refinement, local selection, and reusable histories~\citep{ye2026structured}. SimpleTES also studies trajectory-level post-training, which is distinct from that search baseline. Population search and policy learning are complementary. Exploration in RL~\citep{SuttonRL,weng2020exploration,amin2021survey,yu2024flow} ranges from undirected randomness and entropy regularization~\citep{lillicrap2015continuous,dann2022guarantees,andrychowicz2021matters,yu2026weak,cui2025entropy,shen2025entropy} to directed bonuses based on visitation counts~\citep{lai1987adaptive,li2010contextual,jin2020provably} or prediction error~\citep{pathak2017curiosity,burda2018exploration}, with recent extensions to language models~\citep{bai2025online,gao2025navigate,sun2025curiosity}. In the best-candidate setting of test-time discovery, \method uses an ICWM to predict encoded actor-state transitions and applies its normalized residuals at non-top-$k$ sampled tokens, complementing task-driven policy learning with a selective exploration signal.

\section{Conclusion}
We introduce \method, a test-time learning framework equipped with an ICWM. The ICWM learns token-level transitions in encoded policy hidden states, and its prediction residuals supply a curiosity bonus that supplements task feedback. The bonus is normalized within an epoch, retained at sampled tokens outside the policy's top-$k$ choices, and combined with a token-local task coefficient under an annealed weight. The reported runs improve on the matched task-only control on several mathematical and denoising objectives, while matching the best reported performance on Circle Packing, and exhibit higher code-level structural diversity in the displayed analysis. Prediction error and code-level diversity remain proxies; further controls are needed to establish semantic exploration, the contribution of token selection, and robustness across runs.

\section*{AI use disclosure}
We used generative AI tools to assist with manuscript drafting and editing, LaTeX formatting, literature lookup, checks of mathematical exposition and consistency, and interpretation of reported results. The authors reviewed and edited the AI-assisted material and take full responsibility for the final content.

\bibliography{iclr2027_conference}
\bibliographystyle{iclr2027_conference}

\clearpage
\appendix
\section*{Appendix Contents}
\begingroup
\setcounter{tocdepth}{2}
\hypersetup{linkcolor=black}
\makeatletter
\@starttoc{atoc}
\makeatother
\endgroup
\clearpage
\let\appendixOriginalAddcontentsline\addcontentsline
\renewcommand{\addcontentsline}[3]{%
  \def\appendixRequestedFile{#1}%
  \def\appendixTocFile{toc}%
  \ifx\appendixRequestedFile\appendixTocFile
    \appendixOriginalAddcontentsline{atoc}{#2}{#3}%
  \else
    \appendixOriginalAddcontentsline{#1}{#2}{#3}%
  \fi
}

\newenvironment{promptbox}%
  {\par\addvspace{4pt}\noindent\rule{\linewidth}{1.1pt}\par\nobreak
   \vspace{-6pt}\scriptsize}%
  {\par\vspace{-6pt}\noindent\rule{\linewidth}{1.1pt}\par\addvspace{4pt}}

\section{Implementation Details}
\label{app:impl}

\paragraph{Experimental configuration.}
The configured discovery budget uses $E=50$ epochs, a target of $N=8$ parent groups per epoch, and $G=64$ rollouts per group at temperature $1.0$, corresponding to $25{,}600$ generation slots. \textsc{Best-of-25600} uses that nominal sample budget. Equal generation counts do not imply equal token counts, wall-clock time, or training cost. Archive bootstrap and realized counts are discussed below. We train low-rank adaptation (LoRA) adapters of rank $32$ with Adam at learning rate $4 \times 10^{-5}$ and reference-correction coefficient $\lambda \!=\! 0.1$, and the ICWM (encoder and forward predictor) with AdamW at $5 \times 10^{-6}$ on mini-batches of $8$ rollouts. The exploratory mask of Equation~\eqref{eq:mask} uses $k \!=\! 1$ in every run, and the initial curiosity weight $\eta_0$ is the only task-dependent hyperparameter of \method: following the best settings in Figure~\ref{fig:method_ablation_radar}(a), we use $\eta_0=0.5$ for Erd\H{o}s, Hadamard, and denoising, and $\eta_0=0.3$ for AC1--AC3. Circle Packing retains the default $\eta_0=0.3$.

Table~\ref{tab:hparams} summarizes the reported hyperparameters of \method. The method notation is defined in \S\ref{sec:method}, and the training pseudocode is provided in Appendix~\ref{app:training}. The initial curiosity weight is task-dependent; the other listed settings are shared across tasks and backbone scales.

\begin{table}[h]
\centering
\caption{\textbf{Reported hyperparameters.} The discovery loop and policy optimization follow \S\ref{sec:prelim} and the task
coefficients of \S\ref{sec:task-coefficients}; the ICWM-based curiosity block is specific to \method. Setting $\eta_0 = 0$ removes
the intrinsic reward.}
\label{tab:hparams}
\renewcommand{\arraystretch}{1.15}
\begin{tabular}{ll}
\toprule
Hyperparameter & Value \\
\midrule
\multicolumn{2}{l}{\textit{Discovery loop}} \\
Epochs $E$ & 50 \\
Target parent groups per epoch $N$ & 8 \\
Rollouts per parent $G$ & 64 \\
Configured generation budget $ENG$ & 25,600 \\
PUCT exploration constant $c$ & 1.0 \\
Maximum archive size & 1,000 \\
Retained children per parent & 2 \\
Sampling temperature & 1.0 \\
Context window & 32,768 tokens \\
\midrule
\multicolumn{2}{l}{\textit{Policy optimization}} \\
LoRA rank / scaling factor / dropout & 32 / 32 / 0 \\
LoRA target modules & all attention and MLP projections, unembedding \\
Optimizer & Adam \\
Learning rate & $4 \times 10^{-5}$ \\
$(\beta_1, \beta_2)$, $\epsilon$ & $(0.9, 0.95)$, $10^{-8}$ \\
Weight decay & 0 \\
Substeps per epoch & 1 \\
Reference-correction coefficient $\lambda$ & 0.1 \\   
Constant-reward groups & dropped \\
Optimizer state & persisted across epochs \\
\midrule
\multicolumn{2}{l}{\textit{ICWM and bonus}} \\
Encoder $\phi_\psi$ / forward predictor $f_\psi$ & three-layer MLP with rectified linear units (ReLU) \\
Forward-model intermediate width & backbone hidden size \\
Precision & \texttt{fp32} \\
Optimizer & AdamW \\
Learning rate & $5 \times 10^{-6}$ \\
$(\beta_1, \beta_2)$ & $(0.9, 0.95)$ \\
Mini-batch size & 8 rollouts \\
Exploratory mask $k$ & 1 \\
Annealing floor $\alpha$ & 0.1 \\
Initial curiosity weight $\eta_0$ & 0.5 (Erd\H{o}s, Hadamard, denoising); 0.3 (others) \\
ICWM and optimizer state & checkpointed and reloaded each epoch \\
\midrule
\multicolumn{2}{l}{\textit{Infrastructure}} \\
Generation backend & vLLM via Ray Data \\
Training backend & DeepSpeed \\
Evaluation & task-specific sandbox, isolated from training \\
\bottomrule
\end{tabular}
\end{table}

We expand the rows of Table~\ref{tab:hparams} below. Each epoch of a run executes five stages in a fixed
order (parent sampling, generation, evaluation, archive update, and training), and each stage checkpoints its
output to disk before the next begins, so a run is resumable at epoch granularity after preemption.

\paragraph{Discovery loop.}
A run addresses one problem description $d$. The archive $\mathcal{H}$ is initialized with a seed candidate containing a reference program and its score: a published or hand-written construction for mathematical tasks, and MAGIC for denoising (\S\ref{app:task-denoising}). The configured budget is $E=50$ epochs, a target of $N=8$ parent groups per epoch, and $G=64$ completions per group, giving $ENG=25{,}600$ generation slots. This is a configured budget, not a consequence of the distinct-parent rule below: with only one seed, the archive cannot initially supply eight unrelated parents. An exact realized count requires a specified bootstrap or refill rule and verification against the run logs. Algorithm~\ref{alg:training} therefore operates on the parents actually selected rather than assuming that the requested number is always available. Each request uses a sampling seed derived from its index within the epoch; different seeds do not guarantee different completions.

\paragraph{Archive and parent selection.}
Write $v_p$ for the archive value of candidate $p$ (the task objective, sign-flipped for minimization), $n_p$ for its visit count, and $q_p$ for the best child value observed so far, falling back to $v_p$ if there is no scored child. Let $V$ be the total number of rollouts drawn. The PUCT score is
\begin{equation}
\label{eq:puct}
\mathrm{PUCT}(p)=q_p+c\,\sigma_{\mathcal H}\,\rho_p\frac{\sqrt{1+V}}{1+n_p},
\qquad
\rho_p=\frac{|\mathcal H|-\mathrm{rank}_0(v_p)}{\sum_{q\in\mathcal H}(|\mathcal H|-\mathrm{rank}_0(v_q))}.
\end{equation}
Here $c=1.0$, and $\mathrm{rank}_0$ orders candidates from best to worst starting at zero, with ties resolved by a fixed ordering. The archive spread $\sigma_{\mathcal H}$ is the non-seed value range, floored at $10^{-6}$; if fewer than two non-seed states exist, the floor is used. This makes the score well-defined for a singleton archive. 

A rollout from $p$ increments the visits of $p$ and its ancestors. Parents are chosen greedily by PUCT, removing a selected candidate's ancestors and descendants from contention for that selection round. This rule can return fewer than $N$ parents. Feasible, non-duplicate children are inserted, then the archive retains the two highest-valued children per parent and is truncated to its $1{,}000$ highest-valued candidates, keeping the seed. Insertion does not require a child to improve on its parent.

\paragraph{Task coefficients and temperature calibration.}
Equation~\eqref{eq:loo} defines the leave-one-out task coefficient in the main text. The finite-sample ratio is not generally unbiased. Groups with constant rewards are removed before both actor and ICWM training. For a retained group, temperature calibration targets
\[
\mathrm{KL}\!\left(\mathrm{softmax}(\beta R)\,\|\,\mathrm{unif}_G\right)=\delta,
\qquad \delta=\log 2,
\]
using upper-bracket expansion and $60$ bisection steps when a finite bracket exists. The target is not always attainable: with $m$ tied maxima, the limiting divergence is $\log(G/m)$. A robust calibration routine must check this bound and use a documented finite cap or fallback instead of unbounded bracket expansion. The target and bisection count alone do not specify that boundary behavior. The returned finite $\beta$ is held fixed in the actor update. The token-local reference correction follows Equations~\eqref{eq:ref-logratio}--\eqref{eq:kl}, with the mean taken over trainable tokens in $\mathcal D_\tau$.

\paragraph{ICWM.}
The ICWM of \S\ref{sec:wm-curiosity} consists of the encoder and forward predictor. The encoder $\phi_\psi$ is a three-layer MLP $d_h \to 2d_h \to 2d_h \to d_h$ with ReLU activations, and the forward predictor
$f_\psi$ is a three-layer MLP $2d_h \to d_{\text{int}} \to d_{\text{int}} \to d_h$ that consumes the concatenation of
the encoded state and the action embedding, with $d_h$ the backbone hidden size and $d_{\text{int}} = d_h$ by default.
Both run in \texttt{fp32} regardless of the backbone's compute dtype, since the residual $e_t$ is a difference of
nearby vectors and is the quantity the intrinsic reward is read from. Within an epoch the ICWM is updated on
mini-batches of $8$ rollouts: each rollout's contribution to Equation~\eqref{eq:icmloss} is backpropagated
separately and normalized by the mini-batch's total trainable-token count, so gradients accumulate exactly as they
would for a single batched loss while only one rollout's activation graph is resident at a time. The optimizer
steps once per nonempty trainable mini-batch. The residuals used for policy training are cached and detached before each ICWM step. Equations~\eqref{eq:epoch-stats}--\eqref{eq:whiten} then normalize the cached, masked residuals once over $\mathcal I_\tau$: real completion positions of retained groups, including zero contributions from inserted suffixes and excluding prompts and padding. Residuals come from successive predictor states; they are not recomputed with a common final predictor. This normalization is epoch-wide rather than mini-batch-local. The
ICWM weights and its optimizer state are written alongside the adapter at the end of every epoch and reloaded at
the start of the next, so ICWM training continues across epochs rather than restarting. This persistence does not itself prevent feature collapse or guarantee decreasing prediction error.

\paragraph{Evaluation sandbox.}
Candidate programs are executed in a separate process, under a separate Python environment from the training
stack, with no filesystem or network access and a per-task CPU allocation and wall-clock limit
(Table~\ref{tab:task-overview}). A candidate that raises, returns a malformed object, violates the task's
feasibility constraints, or exceeds its wall-clock limit is assigned a correctness of zero and therefore a reward
of zero; it is still recorded in the epoch's log but is never inserted into the archive. The evaluator's Python
environment is pinned per task and is deliberately not upgraded in step with the training runtime, so that a
runtime dependency bump cannot silently change what a score means. Each rollout's evaluation seed is its global
index, which fixes the evaluator random seed independently of worker assignment. Wall-clock-limited candidate searches can still vary with hardware and scheduling.

\section{Training Procedure Details}
\label{app:training}

Algorithm~\ref{alg:training} gives the full training pseudocode corresponding to \S\ref{sec:joint-training}. At the start of an epoch, $\theta_{\mathrm{old}}$ snapshots the actor. Generation, ICWM feature extraction, and the token-local reference correction all use that snapshot; no actor step occurs until the final policy update. Disabling the adapter yields the initial reference policy $\pi_{\theta_0}$. The base weights remain frozen.

The ICWM mini-batch loop caches each residual before updating $\psi$, and the policy loss subsequently treats both the cached residuals and their epoch statistics as constants. Only the current-policy numerator in Equation~\eqref{eq:pg} receives an actor gradient. Equation~\eqref{eq:icmloss} trains the encoder and predictor but does not backpropagate through actor hidden states or action embeddings. Mini-batches with zero trainable tokens are skipped and have zero cached residuals. If the retained epoch has no trainable tokens, both training updates are skipped while the evaluated candidates and archive changes remain recorded.

Setting $\eta_0=0$ bypasses the curiosity block and sets $b_t=0$, leaving the same task-only actor update. This equivalence concerns the matched TTT-Discover control; it does not assert that ThetaEvolve (test-time RL) shares the same training implementation. AlphaEvolve/OpenEvolve, ShinkaEvolve, Best-of-25600, and the SimpleTES search baseline keep their proposing policies fixed during discovery. Constant-reward filtering applies before the curiosity block, so the method does not provide intrinsic-only updates for those groups.

\begin{algorithm}[!t]
\caption{\textbf{One epoch of \method: rollout collection and training.}}
\label{alg:training}
\small
\begin{algorithmic}[1]
\Require Archive $\mathcal{H}$; actor $\theta$; reference $\theta_0$; ICWM $\psi$; epoch $\tau$
\State $\theta_{\mathrm{old}}\gets\theta$; compute $\eta_\tau$ by Equation~\eqref{eq:eta}
\State Select available parents by the archive rule (Appendix~\ref{app:impl})
\For{each selected parent $p$}
    \State Render $x(d,p)$; sample $G$ completions from $\pi_{\theta_{\mathrm{old}}}$
    \State Record token log-probabilities, $m_t$, and $u_t$ (Equation~\eqref{eq:mask})
    \State Evaluate the extracted programs; record rewards and feasibility
\EndFor
\State Insert feasible, non-duplicate candidates; prune the archive
\State Form $\mathcal{D}_\tau$ by dropping constant-reward groups
\If{$M_\tau>0$}
    \State Obtain each group's calibrated $\beta$; form $A_i^{\mathrm{task}}$ (Equation~\eqref{eq:loo})
    \State Compute reference log-probabilities and $A_t$ (Equations~\eqref{eq:ref-logratio}--\eqref{eq:kl})
    \If{$\eta_0>0$} \label{alg:line-icm-begin}
        \State Initialize cached residuals to zero; extract frozen $h_j$ with a no-gradient forward of $\pi_{\theta_{\mathrm{old}}}$
        \For{each mini-batch $\mathcal{B}\subset\mathcal{D}_\tau$ with $\sum_{y\in\mathcal{B}}\sum_t m_t>0$}
            \State Form transitions and cache detached, pre-update $e_t$ (Equations~\eqref{eq:sa}, \eqref{eq:icm})
            \State Update ICWM parameters $\psi$ using $\mathcal{L}_{\mathrm{ICWM}}$ (Equation~\eqref{eq:icmloss})
        \EndFor
        \State Normalize cached residuals across $\mathcal{I}_\tau$ (Equations~\eqref{eq:epoch-stats}--\eqref{eq:whiten}); form $b_t$ (Equation~\eqref{eq:bonus})
    \Else
        \State $b_t\gets0$ for every completion token
    \EndIf \label{alg:line-icm-end}
    \State Form detached $\tilde A_t$ (Equation~\eqref{eq:final})
    \State Take one adapter optimizer step on $\mathcal{L}_{\mathrm{PG}}$ (Equation~\eqref{eq:pg})
\EndIf
\State Checkpoint the archive, actor, ICWM, and optimizer states
\end{algorithmic}
\end{algorithm}

\section{Inference Procedure}
\label{app:inference}

\paragraph{Chat template.}
Every rollout is a single user turn. We do not inject a system prompt: the task specification and the state
context are the entire user message, which keeps the prompt identical to what the evaluator's notion of the task
implies and avoids a second place where behaviour could be steered. The assistant turn is opened with a
\texttt{<think>} tag so that the backbone enters its reasoning mode deterministically rather than depending on the
model to emit the tag itself. Generation stops at \texttt{<|im\_end|>}. Table~\ref{tab:prompt-template} gives the
rendered form.

\begin{center}\small
\captionof{table}{\textbf{Chat template.} \texttt{\{task\_prompt\}} is replaced by the per-task prompt of
\S\ref{app:prompt}, which already contains the state context. No system message is rendered.}
\label{tab:prompt-template}
\end{center}
\begin{promptbox}
\begin{verbatim}
<|im_start|>user
{task_prompt}<|im_end|>
<|im_start|>assistant
<think>
\end{verbatim}
\end{promptbox}

\paragraph{Sampling.}
Rollouts are sampled at temperature $1.0$ with no top-$p$ or top-$k$ truncation, within a $32{,}768$-token context.
Each request carries a seed equal to the rollout's index in the epoch. This fixes the sampling seed but does not ensure unique completions or bitwise reproducibility across execution backends. We request $k+1$ candidate log-probabilities per position and retain the sampled token's log-probability. This suffices to identify membership in the top-$k$ list under the engine's ordering; at $k=1$ the requested candidate count is two. The exploratory mask $u_t$ is
computed from those returned log-probabilities during generation, so it is available to the trainer without any
recomputation and without a second forward pass.

\paragraph{Two-phase budget forcing.}
A discovery rollout has to finish with an extractable program; a rollout that spends its entire context reasoning
and never emits a code fence scores zero regardless of the quality of the reasoning. We therefore generate in two
phases, applying budget forcing to the thinking span. Phase~1 samples with a cap on the
thinking budget. If the rollout terminates normally, or has already emitted the answer marker
\texttt{```python}, it is kept as is and no second phase runs. If instead phase~1 stopped because it hit its cap
while still inside the think block, we append a forced suffix that closes the think block and opens the answer
fence, and phase~2 resumes from the concatenated token sequence with whatever context remains, less a $50$-token
buffer. A rollout whose remaining budget is already exhausted at that point is recorded with finish reason
\texttt{budget\_exhausted} and evaluated on whatever it produced. Table~\ref{tab:budget-forcing} lists the two
suffixes and the branch that selects them.

Two bookkeeping details matter for the rest of the method. The forced-suffix tokens were written by us rather than
sampled, so they are excluded from the trainable mask ($m_t = 0$) and carry a placeholder log-probability; they
therefore contribute no policy gradient. They are also assigned $u_t = 0$, so they never attract an intrinsic
reward. At sampled positions, the additional non-top-$k$ condition determines whether the curiosity bonus is retained. Phase~2 is resumed from pre-tokenized ids rather than from re-tokenized text, so the boundary
between the two phases cannot perturb the token sequence the trainer later scores.

\begin{center}\small
\captionof{table}{\textbf{Forced suffixes used by two-phase budget forcing.} The branch is decided by token
subsequence matching on the phase-1 output, not by string matching, so it is robust to markers that do not
round-trip through decoding.}
\label{tab:budget-forcing}
\end{center}
\begin{promptbox}
\begin{verbatim}
phase-1 state                                      action
-----------------------------------------------------------------------------
terminated normally (stop sequence emitted)        keep; no phase 2
"```python" already present in phase-1 output      keep; no phase 2
hit cap, still inside <think>                      append "\n</think>\n\n```python\n"
hit cap, "</think>" is the final token             append "\n\n```python\n"
remaining context <= 0 after the suffix            record as budget_exhausted
\end{verbatim}
\end{promptbox}

\paragraph{Answer extraction.}
The evaluator takes the \emph{last} fenced \texttt{```python} block in the completion as the candidate program.
Taking the last rather than the first block is what makes the forced suffix well defined: a rollout that
speculated about several programs while reasoning is scored on the one it committed to at the end. A completion
with no fenced block is scored zero with the diagnostic ``cannot extract python code from model response''.

\section{Task Descriptions}
\label{app:tasks}

Table~\ref{tab:task-overview} summarizes the seven discovery problems. All of them share the same interface: the
policy emits a Python program, a sandbox executes it under a wall-clock limit and a CPU allocation, and a
deterministic scorer returns a scalar. They differ in what the program must return, how the scalar is computed
from it, and how that scalar is mapped to the reward the RL loop consumes.

\begin{table}[h]
\centering
\small
\setlength{\tabcolsep}{4pt}
\renewcommand{\arraystretch}{1.15}
\caption{\textbf{Discovery tasks.} ``Limit'' is the hard wall-clock limit the sandbox enforces per candidate; the
prompt separately advertises a soft compute budget to the model, and where the two differ the hard limit binds.
``Reward'' is the map from the raw objective $r$ to the scalar the RL loop consumes; a candidate that fails the
task's feasibility check receives reward $0$ irrespective of $r$.}
\label{tab:task-overview}
\begin{tabular}{llcccl}
\toprule
Task & Objective & Dir. & Limit & CPUs & Reward from raw score $r$ \\
\midrule
Erd\H{o}s'       & overlap constant $C_5$          & $\downarrow$ & 530\,s  & 1 & $1/(10^{-8} + r)$ \\
AC1              & 1st autocorrelation bound       & $\downarrow$ & 1100\,s & 2 & $1/(10^{-8} + r)$ \\
AC2              & 2nd autocorrelation bound       & $\uparrow$   & 1100\,s & 2 & $r$ \\
AC3              & 3rd autocorrelation bound       & $\downarrow$ & 1100\,s & 2 & $3\,\ell^{3}$, $\ell$ linear in $-r$ \\
Circle Packing   & sum of radii, $n{=}26$          & $\uparrow$   & 530\,s  & 1 & $r$ \\
HadamardMatrix   & $|\det M| / \det_{\mathrm{ref}}$, $n{=}29$ & $\uparrow$ & 700\,s  & 2 & $r$ \\
Denoising        & reconstruction MSE              & $\downarrow$ & 600\,s  & 2 & $1/(10^{-8} + r)$ \\
\bottomrule
\end{tabular}
\end{table}

\subsection{Mathematical discovery}

\textbf{Erd\H{o}s' minimum-overlap problem.} For a step function $h : [0,2] \to [0,1]$ normalized so that
$\int_0^2 h = 1$, the objective is the overlap constant
$C_5 = \max_k \int h(x)\,(1 - h(x+k))\,\mathrm{d}x$, which is to be minimized. The program returns a discretized
$h$ as $n$ samples over $[0,2]$; the scorer computes
$C_5 = \max\big(\mathrm{correlate}(h, 1-h)\big) \cdot \mathrm{d}x$ with $\mathrm{d}x = 2/n$ and rejects any
candidate violating $0 \le h_i \le 1$ or not satisfying $\sum_{i} h_i = n/2$. The prompt states the best published bound
$C_5 \le 0.38092$ and sets $0.38080$ as the target.

\textbf{Autocorrelation inequalities.} Three related problems ask for a sequence of step-function heights
optimizing a ratio of norms of the autoconvolution. \textbf{AC1} minimizes
$2n\,\|f * f\|_\infty / (\sum_i f_i)^2$ over nonnegative sequences, target $1.5030$. \textbf{AC2} maximizes
$\|f * f\|_2^2 / (\|f * f\|_1 \|f * f\|_\infty)$, target $0.97$. \textbf{AC3} minimizes
$2n\,\max|f * f| / (\sum_i f_i)^2$ over sequences that \emph{may take negative values}, target $1.4557$; sign
changes are the mechanism by which the autoconvolution peak is reduced, so unlike AC1 and AC2 the scorer does not
clamp negatives. For all three, the exact scorer is given to the model inside the prompt
(\S\ref{app:prompt-math}), so the policy optimizes against the function it will be graded by and the only
uncertainty is the search. AC1 and AC2 additionally include a short prose summary of the published approach; AC3
includes the Host--Vinuesa window-times-oscillation template. AC3 uses a cubic shaping transform rather than a raw or reciprocal
objective: its scores span a wide range early in a run, so the raw value is linearly mapped from
$[1.4557, 3.2]$ onto $[1,0]$, cubed, and scaled by $3$, which concentrates reward differences near the target
instead of near the initial construction.

\textbf{Circle packing.} Pack $n \!=\! 26$ circles into the unit square maximizing the sum of radii. The program
returns centers and radii; the scorer runs a validator, given verbatim in the prompt, that rejects NaNs, negative
radii, circles leaving $[0,1]^2$, and any overlapping pair, each with a tolerance of $10^{-12}$. The best-known
value is $\approx 2.635983$.

\textbf{Hadamard matrix.} Following the HadamardMatrix task of \citet{wang2025thetaevolve}, produce a $29 \times 29$ matrix with entries in $\{-1,+1\}$ maximizing
$|\det M|$. The score is $|\det M|$ divided by the fixed benchmark normalization constant $\det_{\mathrm{ref}}=2^{28}\cdot 7^{12}\cdot 342$ for $n=29$, matching their Appendix~A.5.
Here ``Hadamard'' is the benchmark label for a maximal-determinant search; an order-29 sign matrix is not a Hadamard matrix with mutually orthogonal rows. The quoted prompt suggests Paley and Sylvester constructions, local search over entry flips, and circulant structure as
starting points.

\subsection{Single-cell denoising}
\label{app:task-denoising}

The policy evolves a Python function \texttt{magic\_denoise(X, **kwargs)} that maps a raw scRNA-seq count matrix
to denoised counts. We follow the OpenProblems dataset protocol used by \citet{yuksekgonul2026learning}: Pancreas supplies discovery feedback, while PBMC and Tabula Muris Senis Lung (Tabula) are held out. Within each dataset, observed molecules are split into input and target counts by binomial sampling. The reported Score is the mean of the benchmark-normalized MSE and Poisson scores, each normalized relative to no-denoising and perfect-denoising references. Two properties of this protocol matter for reading the results.

\textbf{The training signal and the reported metric are different quantities.} During the run, each candidate is
scored on the held-in Pancreas corpus and its reconstruction MSE determines the reward and archive ranking. The numbers reported in
\S\ref{sec:exp} are instead the OpenProblems benchmark Score on two corpora, \textbf{PBMC} and \textbf{Tabula},
neither of which the loop ever observes. Transfer from the training corpus to the held-out ones is not monotone
over a run, so per-epoch held-out curves are computed rather than read off the final epoch.

\textbf{Held-out epochs are selected by the training objective, never by the held-out score.} For each run we
select the epoch by its in-domain MSE and report that epoch's PBMC and Tabula numbers together. Quoting each
corpus at its own best epoch would describe a solution that does not exist and would select on the test set.
Held-out evaluation uses \texttt{random\_state=42}; repeating evaluation of a fixed candidate is distinct from repeating the stochastic discovery run.

\textbf{Reward and the Poisson constraint.} MSE is measured in log-normalized space after total-count
normalization to $10^4$; the reported Poisson metric is the NLL against the held-out molecules. On Pancreas, the normalized Poisson score
acts as a hard gate rather than a reward term: a candidate is accepted only when
$0.97 \le (\,0.257575 - \text{poisson})/(0.257575 - 0.031739\,) \le 1$, following \citet{yuksekgonul2026learning}, and the
reward of an accepted candidate is $1/\text{MSE}$. The training gate is not applied to the held-out PBMC and Tabula results. On held-out data we clamp the normalized Poisson score to $[0,1]$
following the OpenProblems definition and exclude candidates that produce a negative Poisson NLL, which would
otherwise report a normalized score above $1$ without being a valid denoiser. The seed
program is the MAGIC baseline in its reversed-normalization form, which denoises in square-root space before
normalizing; on the two held-out corpora it attains Score $0.6416$ (PBMC) and $0.6414$ (Tabula).

\section{Prompt Usage}
\label{app:prompt}

Each rollout's prompt is assembled from two parts. A \textbf{static task specification} states the problem, the
scorer, the return contract, and the resource limits; it is fixed for the lifetime of a run. A \textbf{dynamic
state context} describes the particular archive node being expanded: the program that node holds, the score it
achieved, how that compares with its own parent, the remaining gap to the target, and the tail of the program's
standard output. The state context is what turns a fixed task description into a step of a search, by carrying previous candidates into the next prompt; policy and ICWM updates also carry experience across epochs. It is spliced into the middle of the
specification, before the final instruction block.

\S\ref{app:prompt-state} gives the state-context template, which is shared by all six mathematical tasks;
\S\ref{app:prompt-math} and \S\ref{app:prompt-denoise} give the per-task specifications that wrap it. In every
listing, \texttt{\{braced\}} names are substituted at render time. Long lines are wrapped for display, and a
handful of non-ASCII glyphs in the original strings (\texttt{Erd\H{o}s}, subscripted constants, arrows and
inequality signs) are transliterated to ASCII here; the strings sent to the model are otherwise unchanged.

\subsection{Shared state context}
\label{app:prompt-state}

Table~\ref{tab:prompt-state} gives the template. Three branches select the value report: a node with a parent gets
the before/after pair, a node without one gets its own value, and the seed node, which has not been scored as a
child of anything, gets only the target. Standard output is included because the tasks invite the model to
instrument its own search; it is truncated to the last $500$ characters (last $3{,}000$ for denoising) with an
explicit truncation marker, so a program that prints in a loop cannot crowd the specification out of the context.
The trailing construction length is emitted for the autocorrelation tasks only, where sequence length is itself a
search dimension.

\begin{center}\small
\captionof{table}{\textbf{State-context template shared by the six mathematical tasks.} \texttt{\{metric\}} and
\texttt{\{direction\}} are per-task constants; the remaining fields come from the archive node being expanded.
Exactly one of the three value branches is rendered.}
\label{tab:prompt-state}
\end{center}
\begin{promptbox}
\begin{verbatim}
You are iteratively optimizing {metric}.

  [if the node carries a program]
Here is the last code we ran:
```python
{parent_code}
```
  [else]
No previous code available.

  [branch 1: node has a parent and a value]
Here is the {metric} before and after running the code above
({direction} is better): {before:.6f} -> {after:.6f}
Target: {target}. Current gap: {gap:.6f}. Further improvements will also be
generously rewarded.

  [branch 2: node has a value but no parent]
Current {metric} ({direction} is better): {after:.6f}
Target: {target}. Current gap: {gap:.6f}. Further improvements will also be
generously rewarded.

  [branch 3: seed node]
Target {metric}: {target}

  [if the program printed anything]

--- Previous Program Output ---

                 ...(TRUNCATED)...
{last 500 characters of stdout}
--- End Output ---

  [AC1, AC2, AC3 only]
Length of the construction: {n}
\end{verbatim}
\end{promptbox}

\subsection{Mathematical discovery prompts}
\label{app:prompt-math}

\paragraph{Erd\H{o}s' minimum-overlap problem.}
The specification states the continuous problem, the discretization the scorer assumes, and the two feasibility
constraints, then names the published record and the target. The closing block is conditional: a node that
already holds a program is told to do something structurally different from it, while the seed node is simply
told to write code. Table~\ref{tab:prompt-erdos} gives the prompt.

\begin{center}\small
\captionof{table}{\textbf{Erd\H{o}s' minimum-overlap prompt.} \texttt{\{budget\_s\}} is the advertised compute
budget, \texttt{\{n\}} the length of the parent construction, and \texttt{\{state\_context\}} the block of
Table~\ref{tab:prompt-state}.}
\label{tab:prompt-erdos}
\end{center}
\begin{promptbox}
\begin{verbatim}
You are an expert in harmonic analysis, numerical optimization, and mathematical
discovery.
Your task is to find an improved upper bound for the Erdos minimum overlap
problem constant C_5.

## Problem

Find a step function h: [0, 2] -> [0, 1] that **minimizes** the overlap integral:

$$C_5 = \max_k \int h(x)(1 - h(x+k)) dx$$

**Constraints**:
1. h(x) in [0, 1] for all x
2. int_0^2 h(x) dx = 1

**Discretization**: Represent h as n_points samples over [0, 2].
With dx = 2.0 / n_points:
- 0 <= h[i] <= 1 for all i
- sum(h) * dx = 1 (equivalently: sum(h) == n_points / 2 exactly)

The evaluation computes: C_5 = max(np.correlate(h, 1-h, mode="full") * dx)

Smaller sequences with less than 1k samples are preferred - they are faster to
optimize and evaluate.

**Lower C_5 values are better** - they provide tighter upper bounds on the Erdos
constant.

## Budget & Resources
- **Time budget**: {budget_s}s for your code to run
- **CPUs**: 1 available

## Rules
- Define `run(seed=42, budget_s={budget_s}, **kwargs)` that returns
  `(h_values, c5_bound, n_points)`
- Use scipy, numpy, cvxpy[CBC,CVXOPT,GLOP,GLPK,GUROBI,MOSEK,PDLP,SCIP,XPRESS,
  ECOS], math
- Make all helper functions top level, no closures or lambdas
- No filesystem or network IO
- `evaluate_erdos_solution()` and `initial_h_values` (an initial construction,
  if available) are pre-imported
- Your function must complete within budget_s seconds and return the best
  solution found

**Lower is better**. Current record: C_5 <= 0.38092. Our goal is to find a
construction that shows C_5 <= 0.38080.

{state_context}

  [if a parent construction is available]
You may want to start your search from the current construction, which you can
access through the `initial_h_values` global variable (n={n} samples).
You are encouraged to explore solutions that use other starting points to
prevent getting stuck in a local optimum.

  [if the node carries a program]
Reason about how you could further improve this construction.
Ideally, try to do something different than the above algorithm. Could be using
different algorithmic ideas, adjusting your heuristics, adjusting / sweeping
your hyperparemeters, etc.
Unless you make a meaningful improvement, you will not be rewarded.

  [else]
Write code to optimize this construction.
\end{verbatim}
\end{promptbox}

\paragraph{First autocorrelation inequality.}
AC1 hands the model the scorer verbatim, so the objective is unambiguous, together with a prose restatement of the
published linear-programming approach offered explicitly as inspiration rather than as a recipe.
Table~\ref{tab:prompt-ac1} gives the prompt.

\begin{center}\small
\captionof{table}{\textbf{AC1 prompt.} The embedded \texttt{evaluate\_sequence} is the exact scorer used by the
sandbox.}
\label{tab:prompt-ac1}
\end{center}
\begin{promptbox}
\begin{verbatim}
Act as an expert software developer and inequality specialist specializing in
creating step functions with certain properties.

Your task is to generate the sequence of non-negative heights of a step
function, that minimizes the following evaluation function:

```python
import numpy as np

def evaluate_sequence(sequence: list[float]) -> float:
    """
    Evaluates a sequence of coefficients with enhanced security checks.
    Returns np.inf if the input is invalid.
    """
    # --- Security Checks ---

    # Verify that the input is a list
    if not isinstance(sequence, list):
        return np.inf

    # Reject empty lists
    if not sequence:
        return np.inf

    # Check each element in the list for validity
    for x in sequence:
        # Reject boolean types (as they are a subclass of int) and
        # any other non-integer/non-float types (like strings or complex
        # numbers).
        if isinstance(x, bool) or not isinstance(x, (int, float)):
            return np.inf

        # Reject Not-a-Number (NaN) and infinity values.
        if np.isnan(x) or np.isinf(x):
            return np.inf

    # Convert all elements to float for consistency
    sequence = [float(x) for x in sequence]

    # Protect against negative numbers
    sequence = [max(0, x) for x in sequence]

    # Protect against numbers that are too large
    sequence = [min(1000.0, x) for x in sequence]

    n = len(sequence)
    b_sequence = np.convolve(sequence, sequence)
    max_b = max(b_sequence)
    sum_a = np.sum(sequence)

    # Protect against the case where the sum is too close to zero
    if sum_a < 0.01:
        return np.inf

    return float(2 * n * max_b / (sum_a**2))
```

A previous state of the art used the following approach. You can use it as
inspiration, but you are not required to use it, and you are encouraged to
explore.
```latex
Starting from a nonnegative step function $f=(a_0,\dots,a_{n-1})$ normalized so
that $\sum_j a_j=\sqrt{2n}$, set $M=\|f*f\|_\infty$. Next compute
$g_0=(b_0,\dots,b_{n-1})$ by solving a linear program, i.e.\ maximizing
$\sum_j b_j$ subject to $b_j\ge0$ and $\|f*g_0\|_\infty\le M$; as is standard,
the optimum is attained at an extreme point determined by an active set of
binding inequalities, here corresponding to important constraints where the
convolution bound $(f*g_0)(x)\le M$ is tight and limiting. Rescale $g_0$ to
match the normalization, $g=\frac{\sqrt{2n}}{\sum_j b_j}g_0$, and update
$f\leftarrow (1-t)f+t g$ for a small $t>0$. Repeating this step produces a
sequence with nonincreasing $\|f*f\|_\infty$, and the iteration is continued
until it stabilizes.
```

Your task is to write a search function that searches for the best sequence of
coefficients. Your function will have {budget_s} seconds to run, and after that
it has to have returned the best sequence it found. If after {budget_s} seconds
it has not returned anything, it will be terminated with negative infinity
points. All numbers in your sequence have to be positive or zero. Larger
sequences with {budget_s}s of items often have better attack surface, but too
large sequences with 100s of thousands of items may be too slow to search.

You may code up any search method you want, and you are allowed to call the
evaluate_sequence() function as many times as you want. You have access to it,
you don't need to code up the evaluate_sequence() function.

{state_context}

You may want to start your search from one of the constructions we have found so
far, which you can access through the 'height_sequence_1' global variable.
However, you are encouraged to explore solutions that use other starting points
to prevent getting stuck in a local minimum.

Reason about how you could further improve this construction.
Ideally, try to do something different than the above algorithm. Could be using
different algorithmic ideas, adjusting your heuristics, adjusting / sweeping
your hyperparemeters, etc.
Unless you make a meaningful improvement, you will not be rewarded.

Rules:
- You must define the `propose_candidate` function as this is what will be
  invoked.
- You can use scientific libraries like scipy, numpy, cvxpy[CBC,CVXOPT,GLOP,
  GLPK,GUROBI,MOSEK,PDLP,SCIP,XPRESS,ECOS], math.
- You can use up to 2 CPUs.
- Make all helper functions top level and have no closures from function
  nesting. Don't use any lambda functions.
- No filesystem or network IO.
- Do not import evaluate_sequence yourself. Assume it will already be imported
  and can be directly invoked.
- **Print statements**: Use `print()` to log progress, intermediate bounds,
  timing info, etc. Your output will be shown back to you.
- Include a short docstring at the top summarizing your algorithm.

Make sure to think and return the final program between ```python and ```.
\end{verbatim}
\end{promptbox}

\paragraph{Second autocorrelation inequality.}
AC2 maximizes rather than minimizes, and its scorer integrates the autoconvolution numerically, so the embedded
function is longer than AC1's. Its literature note describes a coarse-to-fine scheme (stochastic global search,
projected gradient ascent, then upsampling by entry repetition) rather than a linear program.
Table~\ref{tab:prompt-ac2} gives the prompt.

\begin{center}\small
\captionof{table}{\textbf{AC2 prompt.} Structure matches AC1; the scorer, the optimization direction, the
literature note, and the required entry-point name differ.}
\label{tab:prompt-ac2}
\end{center}
\begin{promptbox}
\begin{verbatim}
Act as an expert software developer and inequality specialist specializing in
creating step functions with certain properties.

Your task is to generate the sequence of non-negative heights of a step
functions, that maximizes the following evaluation function:

```python

import numpy as np

def evaluate_sequence(sequence: list[float]) -> float:
    # Verify that the input is a list
    if not isinstance(sequence, list):
        raise ValueError("Invalid sequence type")

    # Reject empty lists
    if not sequence:
        raise ValueError("Empty sequence")

    # Check each element in the list for validity
    for x in sequence:
        # Reject boolean types (as they are a subclass of int) and
        # any other non-integer/non-float types (like strings or complex
        # numbers).
        if isinstance(x, bool) or not isinstance(x, (int, float)):
            raise ValueError("Invalid sequence element type")

        # Reject Not-a-Number (NaN) and infinity values.
        if np.isnan(x) or np.isinf(x):
            raise ValueError("Invalid sequence element value")

    # Convert all elements to float for consistency
    sequence = [float(x) for x in sequence]

    # Protect against negative numbers
    sequence = [max(0, x) for x in sequence]

    # Check if sum of sequence will be too close to zero
    if np.sum(sequence) < 0.01:
        raise ValueError("Sum of sequence is too close to zero.")

    # Protect against numbers that are too large
    sequence = [min(1000.0, x) for x in sequence]

    convolution_2 = np.convolve(sequence, sequence)
    # --- Security Checks ---

    # Calculate the 2-norm squared: ||f*f||_2^2
    num_points = len(convolution_2)
    x_points = np.linspace(-0.5, 0.5, num_points + 2)
    x_intervals = np.diff(x_points) # Width of each interval
    y_points = np.concatenate(([0], convolution_2, [0]))
    l2_norm_squared = 0.0
    for i in range(len(convolution_2) + 1):  # Iterate through intervals
        y1 = y_points[i]
        y2 = y_points[i+1]
        h = x_intervals[i]
        # Integral of (mx + c)^2 = h/3 * (y1^2 + y1*y2 + y2^2) where
        # m = (y2-y1)/h, c = y1 - m*x1, interval is [x1, x2], y1 = mx1+c,
        # y2=mx2+c
        interval_l2_squared = (h / 3) * (y1**2 + y1 * y2 + y2**2)
        l2_norm_squared += interval_l2_squared

    # Calculate the 1-norm: ||f*f||_1
    norm_1 = np.sum(np.abs(convolution_2)) / (len(convolution_2) + 1)

    # Calculate the infinity-norm: ||f*f||_inf
    norm_inf = np.max(np.abs(convolution_2))
    C_lower_bound = l2_norm_squared / (norm_1 * norm_inf)
    return C_lower_bound

```

A previous state of the art used the following approach. You can use it as
inspiration, but you are not required to use it, and you are encoraged to
explore.
```latex
Their procedure is a coarse-to-fine optimization of the score. It starts with a
stochastic global search that repeatedly perturbs the current best candidate and
keeps the perturbation whenever it improves (Q), with the perturbation scale
gradually reduced over time. Once a good basin is found, they switch to a
deterministic local improvement step, performing projected gradient ascent (move
in the gradient direction and project back to the feasible region). To reach
higher resolution, they lift a good low-resolution solution to a
higher-dimensional one by simply repeating its entries and then rerun the local
refinement. Iterating this explore-refine-upscale cycle yields their final
high-resolution maximizer and the improved lower bound.
```

Your task is to write a search function, construct_function(), that searches for
the best sequence of coefficients. Your function will have {budget_s} seconds to
run, and after that it has to have returned the best sequence it found. If after
{budget_s} seconds it has not returned anything, it will be terminated with
negative infinity points. All numbers in your sequence have to be positive or
zero. Larger sequences with {budget_s}s of items often have better attack
surface, but too large sequences with 100s of thousands of items may be too slow
to search.

You may code up any search method you want, and you are allowed to call the
evaluate_sequence() function as many times as you want. You have access to it,
you don't need to code up the evaluate_sequence() function.

{state_context}

You may want to start your search from one of the constructions we have found so
far, which you can access through the 'height_sequence_1' global variable.
However, you are encouraged to explore solutions that use other starting points
to prevent getting stuck in a local minimum.

Reason about how you could further improve this construction.
Ideally, try to do something different than the above algorithm. Could be using
different algorithmic ideas, adjusting your heuristics, adjusting / sweeping
your hyperparemeters, etc.
Unless you make a meaningful improvement, you will not be rewarded, if you are
stuck you should think about how to get unstuck.

Rules:
- You must define the `construct_function` function as this is what will be
  invoked.
- You can use scientific libraries like scipy, numpy, cvxpy[CBC,CVXOPT,GLOP,
  GLPK,GUROBI,MOSEK,PDLP,SCIP,XPRESS,ECOS], math.
- You can use up to 2 CPUs.
- Make all helper functions top level and have no closures from function
  nesting. Don't use any lambda functions.
- No filesystem or network IO.
- Do not import evaluate_sequence yourself. Assume it will already be imported
  and can be directly invoked. Do not import height_sequence_1 yourself; it will
  already be available.
- **Print statements**: Use `print()` to log progress, intermediate bounds,
  timing info, etc. Your output will be shown back to you.
- Include a short docstring at the top summarizing your algorithm.

Make sure to think and return the final program between ```python and ```.
\end{verbatim}
\end{promptbox}

\paragraph{Third autocorrelation inequality.}
AC3 is the only task whose feasible set includes negative values, and the prompt says so three times (in the
scorer's docstring, in the literature note, and in the construction guidelines) because clamping to
nonnegative values is the natural habit carried over from AC1 and AC2 and it removes the mechanism that makes the
problem tractable. Table~\ref{tab:prompt-ac3} gives the prompt.

\begin{center}\small
\captionof{table}{\textbf{AC3 prompt.} \texttt{\{target\}} is $1.4557$.}
\label{tab:prompt-ac3}
\end{center}
\begin{promptbox}
\begin{verbatim}
You are an expert in computational optimization and harmonic analysis. Your task
is to design a Python program that constructs a discrete function `f` on the
domain [-1/4, 1/4] to minimize the third autocorrelation constant C3, aiming to
beat the SOTA of {target}.

C3 is computed by the following evaluation function (lower is better):

```python
import numpy as np

def evaluate_sequence(sequence: list[float]) -> float:
    """
    Evaluates a discrete function f for the THIRD autocorrelation inequality.
    Computes C3 = abs(2 * n * max(|conv(f, f)|) / (sum(f))**2), lower is better.

    NOTE: function values MAY be negative. Sign changes (phase cancellation)
    are a key mechanism for reducing the autoconvolution peak. Only NaN/inf,
    extreme magnitudes (|x| > 1e10), and a near-zero sum are rejected.
    """
    arr = np.asarray([float(x) for x in sequence], dtype=float)
    if not np.all(np.isfinite(arr)):
        return np.inf
    if np.max(np.abs(arr)) > 1e10:
        return np.inf
    total = float(np.sum(arr))
    if abs(total) < 1e-10:
        return np.inf
    conv = np.convolve(arr, arr)
    max_conv_abs = float(np.max(np.abs(conv)))
    return float(abs(2 * len(arr) * max_conv_abs / (total ** 2)))
```

**Key Insight from Mathematical Literature (Host, Vinuesa):**
The best-known constructions are based on the product of a smooth, oscillating
function and a window function with compact support. A highly successful
continuous analog is `f(x) = (1 + cos(2*pi*x))` for `x` in `[-1/4, 1/4]` and `0`
otherwise. IMPORTANT: function values MAY be negative - sign changes (phase
cancellation) are a key mechanism for reducing the autoconvolution peak. Do NOT
clamp your function to be non-negative.

**Construction Guidelines:**
1.  **Window Function:** Define a "window"/"support" for your function, centered
    and occupying a fraction of the total domain (e.g. the middle 50%, like
    `n//4` to `3*n//4`). The function is zero outside this window.
2.  **Oscillatory Component:** Inside the window use a smooth, symmetric,
    oscillating pattern. `A * (1 + B * cos(C * x))` is a powerful template (with
    B>1 it takes negative values).
3.  **Parameterization:** Explore `support_width`, amplitude `A`, modulation
    `B`, frequency `C`.
4.  **Discretization:** Map the continuous form onto a discrete grid (a length
    around 400-800 works well; n~600 is a good default). Mind boundary
    conditions at the window edges.

**Refinement strategy (to push past SOTA):**
- Perform many local-search iterations (e.g. 5,000-20,000) on the current
  candidate.
- Multi-stage adaptive step size: start with larger changes (+/-0.05), then
  shrink (+/-0.01, +/-0.001) to fine-tune.
- Targeted search: find the indices where `conv(f, f)` has the highest absolute
  values and focus perturbations there - they dominate C3.
- Escape local minima: use a simulated-annealing-style schedule; if stagnant for
  hundreds of iterations, apply a larger random perturbation.
- Sign flipping: systematically test flipping the signs of small segments -
  phase cancellation is the key mechanism for lowering the convolution peak.

Your function will have {budget_s} seconds to run and must return the best
sequence (a Python list of floats, negatives allowed) it found. You may call
evaluate_sequence() as many times as you want - it is already available, do not
redefine or import it.

{state_context}

You may start your search from one of the constructions found so far via the
`height_sequence_1` global variable, but you are encouraged to explore other
starting points to avoid local minima.

Reason about how to further improve the construction. Try something different
from the previous algorithm (different structure, heuristics, or
hyperparameters). Unless you make a meaningful improvement, you will not be
rewarded.

Rules:
- You must define the `propose_candidate` function as this is what will be
  invoked.
- You can use scientific libraries like scipy, numpy, cvxpy[CBC,CVXOPT,GLOP,
  GLPK,GUROBI,MOSEK,PDLP,SCIP,XPRESS,ECOS], math.
- You can use up to 2 CPUs.
- Make all helper functions top level and have no closures from function
  nesting. Don't use any lambda functions.
- No filesystem or network IO.
- Do not import evaluate_sequence yourself. Assume it will already be imported
  and can be directly invoked.
- **Print statements**: Use `print()` to log progress, intermediate bounds,
  timing info, etc. Your output will be shown back to you.
- Include a short docstring at the top summarizing your algorithm.

Make sure to think and return the final program between ```python and ```.
\end{verbatim}
\end{promptbox}

\paragraph{Circle packing.}
The circle-packing prompt is the shortest of the six because the problem needs no restatement; what it does
include is the validator's source, extracted at render time from the evaluator module so that the text in the
prompt cannot drift from the code that runs. This is also the one task whose closing instruction asks for an
explicit plan, in \texttt{<strategy>} tags, before the program. Table~\ref{tab:prompt-circle} gives the prompt.

\begin{center}\small
\captionof{table}{\textbf{Circle-packing prompt.} \texttt{\{n\}} is the number of circles ($26$ in all reported
runs). The validator is inlined from the evaluator's source at render time.}
\label{tab:prompt-circle}
\end{center}
\begin{promptbox}
\begin{verbatim}
You are an expert mathematician specializing in circle packing problems and
computational geometry.

Your task is to pack {n} circles in a unit square [0,1]x[0,1] to maximize the
sum of radii.

We will run the below validation function (read-only, do not modify this):
```python
def validate_packing(centers, radii):
    """
    Validate that circles don't overlap and are inside the unit square

    Args:
        centers: np.array of shape (n, 2) with (x, y) coordinates
        radii: np.array of shape (n) with radius of each circle

    Returns:
        True if valid, False otherwise
    """
    n = centers.shape[0]

    # Check for NaN values
    if np.isnan(centers).any():
        print("NaN values detected in circle centers")
        return False

    if np.isnan(radii).any():
        print("NaN values detected in circle radii")
        return False

    # Check if radii are nonnegative and not nan
    for i in range(n):
        if radii[i] < 0:
            print(f"Circle {i} has negative radius {radii[i]}")
            return False
        elif np.isnan(radii[i]):
            print(f"Circle {i} has nan radius")
            return False

    # Check if circles are inside the unit square
    for i in range(n):
        x, y = centers[i]
        r = radii[i]
        if x - r < -1e-12 or x + r > 1 + 1e-12 or y - r < -1e-12 \
                or y + r > 1 + 1e-12:
            print(f"Circle {i} at ({x}, {y}) with radius {r} is outside the "
                  f"unit square")
            return False

    # Check for overlaps
    for i in range(n):
        for j in range(i + 1, n):
            dist = np.sqrt(np.sum((centers[i] - centers[j]) ** 2))
            if dist < radii[i] + radii[j] - 1e-12:  # Allow for tiny errors
                print(f"Circles {i} and {j} overlap: dist={dist}, "
                      f"r1+r2={radii[i]+radii[j]}")
                return False

    return True
```

{state_context}

Reason about how you could further improve this packing. Consider:
- Are circles placed optimally near boundaries and corners?
- Could a different arrangement (hexagonal, nested, hybrid) yield better
  results?
- Are there gaps that could be filled with repositioned or resized circles?
- Could optimization parameters or methods be improved?

Rules:
- You must define the run_packing function:
  def run_packing() -> tuple[np.ndarray, np.ndarray, float]
- Returns (centers, radii, sum_radii) where centers has shape ({n}, 2) and radii
  has shape ({n},).
- You can use scientific libraries like scipy, numpy, cvxpy, math.
- Centers must lie within [0,1]^2 and radii must be nonnegative.
- The pair (centers, radii) must satisfy non-overlap and boundary constraints.
- Make all helper functions top level and have no closures from function
  nesting. Don't use any lambda functions.
- No filesystem or network IO.
- You need to get really creative and think from first principles.

Make sure to /think step by step, first give your strategy between <strategy>
and </strategy> tags, then finally return the final program between ```python
and ```.
\end{verbatim}
\end{promptbox}

\paragraph{Hadamard matrix.}
The Hadamard prompt states the normalization explicitly (the reported score is $|\det M|$ over the fixed benchmark normalization constant for $n=29$, not the determinant itself) and lists construction families rather than an algorithm.
Table~\ref{tab:prompt-hadamard} reproduces the original prompt, including its informal use of ``Hadamard matrix'' and ``theoretical maximum''; these phrases do not establish an attainable optimum.

\begin{center}\small
\captionof{table}{\textbf{Hadamard matrix prompt.}}
\label{tab:prompt-hadamard}
\end{center}
\begin{promptbox}
\begin{verbatim}
You are an expert mathematician specializing in combinatorial optimization and
Hadamard matrices.

A Hadamard matrix is an n x n matrix with entries +1 or -1 such that the
absolute value of the determinant is maximized. For n=29, the theoretical
maximum determinant is 1270698346568170340352 (this is (2^28) * (7^12) * 342).

Your task is to write a Python function `generate_hadamard_matrix(n)` that
returns a 29x29 matrix with entries +1 or -1 that maximizes the absolute
determinant. The score is |det(M)| / theoretical_max, where higher is better.

Key approaches to consider:
- Start from known Hadamard matrix constructions (Paley, Sylvester, etc.)
- Use local search: flip entries and check if determinant improves
- Use simulated annealing or genetic algorithms
- Exploit mathematical structure (circulant matrices, quadratic residues)
- For n=29, Paley construction using quadratic residues mod 29 is a strong
  starting point

Your function will have 600 seconds to run. The function should return a numpy
array of shape (29, 29) with entries +1 or -1.

{state_context}

Rules:
- You must define the `generate_hadamard_matrix(n)` function as this is what
  will be invoked.
- You can use scientific libraries like scipy, numpy, math.
- You can use up to 2 CPUs.
- Make all helper functions top level and have no closures from function
  nesting. Don't use any lambda functions.
- No filesystem or network IO.
- **Print statements**: Use `print()` to log progress, intermediate bounds,
  timing info, etc. Your output will be shown back to you.
- Include a short docstring at the top summarizing your algorithm.

Make sure to think and return the final program between ```python and ```.
\end{verbatim}
\end{promptbox}

\subsection{Single-cell denoising prompt}
\label{app:prompt-denoise}

The denoising prompt includes domain-specific guidance beyond the problem statement. The closing
``Key Insights'' block records four empirical facts about the benchmark, the most important being that denoising must
happen in raw or square-root space \emph{before} normalization, which is the difference between a normalized
Poisson deviance of ${\approx}0.98$ and ${\approx}0.55$ and therefore between passing and failing the hard gate.
Without it, essentially every early candidate is rejected on the constraint and the run receives no gradient at
all. The prompt also differs from the mathematical tasks in that its state context reports two metrics rather than
one, and it retains a longer tail of standard output ($3{,}000$ characters), since candidates here routinely print
per-stage diagnostics. Table~\ref{tab:prompt-denoise} gives the prompt.

\begin{center}\small
\captionof{table}{\textbf{Single-cell denoising prompt.} The two evaluation functions are the exact scorers.
\texttt{\{mse\}}, \texttt{\{poisson\}}, \texttt{\{norm\}} and \texttt{\{gap\}} are the parent node's measurements;
the seed node reports the MAGIC baseline at MSE $0.2316$.}
\label{tab:prompt-denoise}
\end{center}
\begin{promptbox}
\begin{verbatim}
You are an expert in computational biology and single-cell RNA-seq analysis.
Your task is to develop a denoising algorithm for scRNA-seq count data.

## Problem

Single-cell RNA-seq data is noisy due to technical dropout and low capture
efficiency.
Given noisy count data, predict the true expression levels.

Your prediction is evaluated against held-out molecules using two metrics:
1. **MSE** - Mean Squared Error in log-normalized space (lower is better, this
   is the reward)
2. **Poisson Loss** - Poisson negative log-likelihood (hard constraint)

## Data Format

- Input `X`: numpy array of shape (n_cells, n_genes) - raw count data
- Output: numpy array of same shape - your denoised counts

## Evaluation

Your output is evaluated using these exact functions:

```python
def evaluate_mse(test_data, denoised):
    test_X = scprep.utils.toarray(test_data).copy()
    denoised_X = np.asarray(denoised).copy()
    test_adata = anndata.AnnData(X=test_X)
    denoised_adata = anndata.AnnData(X=denoised_X)
    sc.pp.normalize_total(test_adata, target_sum=10000)
    sc.pp.log1p(test_adata)
    sc.pp.normalize_total(denoised_adata, target_sum=10000)
    sc.pp.log1p(denoised_adata)
    return sklearn.metrics.mean_squared_error(test_adata.X, denoised_adata.X)
```

```python
def evaluate_poisson(train_data, test_data, denoised):
    test_X = scprep.utils.toarray(test_data)
    denoised_X = np.asarray(denoised).copy()
    initial_sum = train_data.sum()
    target_sum = test_X.sum()
    denoised_scaled = denoised_X * target_sum / initial_sum
    return poisson_nll_loss(test_X, denoised_scaled)
```

## Scoring

**Poisson is a HARD CONSTRAINT.** Your solution is REJECTED if
`poisson_norm < 0.97`.
- `poisson_norm = (0.257575 - poisson) / (0.257575 - 0.031739)`
- MAGIC baseline achieves poisson_norm ~ 0.97

**Reward = 1/MSE** (after passing Poisson constraint). Lower MSE = higher
reward.

## Budget & Resources

- **Time budget**: 400s for your code to run
- **CPUs**: 2 available

## Function Signature

```python
def magic_denoise(X, **kwargs):
    # kwargs may include: budget_s, random_state, knn, t, n_pca, solver, decay,
    #                     knn_max, n_jobs
    # Your implementation here
    return denoised_X  # same shape as X
```

## Rules

- Implement `magic_denoise(X, ...)` that returns a denoised array of the same
  shape
- Available libraries: numpy, scipy, sklearn, graphtools, scprep, scanpy,
  anndata
- Make all helper functions top-level (no closures or lambdas)
- No filesystem or network IO

## Key Insights from Benchmarks

- NORMALIZATION ORDER MATTERS: Denoise in raw/sqrt space first, then normalize.
  "Reversed normalization order" achieves Poisson ~0.98 vs ~0.55 for standard
  order.
- Square root transform is variance-stabilizing for Poisson distributions
- Poisson loss is highly affected by low non-zero values - push values < 1
  toward zero
- The original MAGIC with reversed normalization achieves Poisson ~ 0.97 (the
  constraint threshold)


You are iteratively improving a scRNA-seq denoising algorithm (minimize MSE).

Here is the current implementation:
```python
{parent_code}
```

Current metrics (lower is better): MSE={mse:.6f}, Poisson={poisson:.6f}
(norm={norm:.4f})
Target MSE: 0.2. Current gap: {gap:.6f}. Further improvements will also be
generously rewarded.

  [if the program printed anything]
Output from last run:
```
{last 3000 characters of stdout}
```

Write your improved `magic_denoise` function.
\end{verbatim}
\end{promptbox}

\section{Code-Level Solution Diversity Metric}
\label{app:code-diversity}

\paragraph{Pairwise code comparison.}
Figure~\ref{fig:code_solution_diversity} uses OpenEvolve's \textit{island code diversity metric}~\citep{openevolve}. For code strings $c_i,c_j$, its structural dissimilarity is
\begin{equation}
\label{eq:code-diversity-pair}
\begin{aligned}
d_{\mathrm{code}}(c_i,c_j)
={}&0.1\,\big|L(c_i)-L(c_j)\big|
 +10\,\big|N(c_i)-N(c_j)\big|\\
 &+0.5\,\big|S(c_i)\mathbin{\triangle}S(c_j)\big|.
\end{aligned}
\end{equation}
Here $L(c)$ is the number of characters in the code string, $N(c)$ counts newline characters, and $S(c)$ is the set of distinct characters. The symmetric difference $S(c_i)\mathbin{\triangle}S(c_j)$ contains character types appearing in exactly one string. OpenEvolve implements the line-count term with \texttt{code.count("\textbackslash n")}, rather than counting nonempty lines.

\paragraph{Aggregation.}
Let $\mathcal{Q}_\tau$ denote the selected unordered program pairs at epoch $\tau$. The island statistic is their mean dissimilarity:
\begin{equation}
\label{eq:code-diversity-island}
D_\tau=\frac{1}{|\mathcal{Q}_\tau|}
\sum_{(i,j)\in\mathcal{Q}_\tau}d_{\mathrm{code}}(c_i,c_j),
\end{equation}
with $D_\tau=0$ when no pair is available. OpenEvolve's island routine sorts programs by identifier, considers up to five programs, and averages up to six pairwise comparisons. Its separate candidate-level routine averages $d_{\mathrm{code}}(c,r)$ over reference programs $r\in\mathcal{R}$ whose code differs from $c$, returning zero for an empty comparison set. These are different aggregations of the same pairwise heuristic.\footnote{Source: \url{https://github.com/algorithmicsuperintelligence/openevolve/blob/main/openevolve/database.py}, routines \texttt{\_fast\_code\_diversity}, \texttt{\_calculate\_island\_diversity}, and \texttt{\_get\_cached\_diversity}.}

\paragraph{Interpretation.}
The score is nonnegative and unnormalized; larger values indicate greater differences in these code features. It is not an execution-based or semantic distance: formatting and comments can change it without changing behavior, while different programs can share the same length, newline count, and character set and receive zero dissimilarity. Accordingly, we interpret Figure~\ref{fig:code_solution_diversity} as code-level structural variation and compare methods within each task. Figure~\ref{fig:modality_routing} reports stage-level summaries of this code-diversity diagnostic on Erd\H{o}s and Circle Packing: early (epochs 0--12), middle (13--25), late (26--38), and final (39--49). Larger displayed values indicate greater structural variation, consistent with the epoch-level trajectories in Figure~\ref{fig:code_solution_diversity}.

\end{document}